USING UAS IMAGERY AND COMPUTER VISION TO SUPPORT SITE-SPECIFIC WEED

CONTROL IN CORN

A Thesis
Submitted to the Graduate Faculty
of the
North Dakota State University
of Agriculture and Applied Science

By

Ranjan Sapkota

In Partial Fulfillment of the Requirements
for the Degree of
MASTER OF SCIENCE

Major Department:
Agricultural and Biosystems Engineering

December 2021

Fargo, North Dakota

# North Dakota State University
## Graduate School

**Title**

USING UAV IMAGERY AND COMPUTER VISION TO SUPPORT SITE-SPECIFIC WEED CONTROL IN CORN

**By**

Ranjan Sapkota

The Supervisory Committee certifies that this *disquisition* complies with North Dakota State University's regulations and meets the accepted standards for the degree of

**MASTER OF SCIENCE**

SUPERVISORY COMMITTEE:

Dr. Paulo Flores
Chair

Dr. Michael Ostlie

Dr. Zhao Zhang

Dr. Ravi Yellavajjala

Approved:

| March 14, 2022 | Dr. Leon Schumacher |
|---|---|
| Date | Department Chair |

# ABSTRACT

Currently, a blanket application of herbicides across the field without considering the spatial distribution of weeds is the most used method to control weeds in corn. Unmanned aerial systems (UASs) can provide high spatial resolution imagery, which can be used to map weeds across a field with a high spatial and temporal resolution during early growing season to support site-specific weed control (SSWC). The proposed approach assumes that plants growing outside the corn rows are weeds that need to be controlled. For that, we are proposing the use of "Pixel Intensity Projection" (PIP) algorithm for the detection of corn rows on UAS imagery. After being identified, corn rows were then removed from the imagery and the remaining vegetation fraction was assumed to be weeds. A weed prescription map based on the remaining vegetation fraction was created and implemented through a commercial sprayer field weed control.

# ACKNOWLEDGMENTS

First of all, I would like to thank my advisor, Dr. Paulo Flores, Assistant Professor, Department of Agricultural and Biosystems Engineering (ABEN), North Dakota State University (NDSU), for his enormous support, guidance, encouragement, and his continuous dedication in assisting me throughout graduate school. He made an environment for me to grow as a researcher in the field of precision agriculture and the experiences he presented to me during my masters were beyond my expectations of any professor. I could not imagine finishing my master's degree without his time and help, I greatly appreciate it.

I would like to thank my committee members Dr. Michael Ostlie, Dr. Zhao Zhang, and Dr. Ravi Kiran Yellavajjala for their guidance, suggestions and encouragement throughout my research.

I want to acknowledge the North Dakota Corn Council (NDCC) for funding this research project, as well as the NDSU Agricultural Experiment Station for additional funding of the project. In addition, I want to acknowledge the Carrington Research Extension Center (CREC) for providing research trial field. I would especially like to thank Richard Richter and Chad Richter for their great support during machine operation. Many thanks as well to the folks at the High Plains Equipment- CaseIH Dealership in Carrington, for their help getting the sprayer set up with the hardware and software needed for my study. In addition, I would like to thank Diego Gris, Jithin Matthew, Anup Kumar Das for helping me during the research data collection.

Many thanks to Dr. Zhulu Lin for teaching an amazing course on numerical system modeling, which helped me a lot to think big picture of my programming skill. I want to thank Dr. Stephanie Day for providing wonderful content about basic ArcGIS. Special Thanks to Dr. David Kramar for assisting me with GIS-related problems during the research.

I would like to thank Jana Daeuber for her technical assistance related to software installation and configuration during the research. I would explicitly mention Kristina Caton for her support and consultation during my thesis writing.

Most of all, I would like to thank my parents for their continuous encouragement throughout my graduate school. I would explicitly mention Anil Karmacharya, Nabin Karki, Anuj Ghimire, Girish Raj Uprety, Rashmi Satyal, Kamal Upadhyay, and Shirjana Bhattarai for their emotional support during my master's studies. Special thanks to Uday Bhanu Prakash, Ph.D. candidate, for motivating me during my research. Many thanks to Md. Sanaul Huda, Sagar Regmi, IBK Ajayi-Banji and Maisha Rehnuma, who supported me and motivated me throughout my masters.

Finally, I would like to thank all faculty and staff of ABEN, NDSU for being so nice and supportive. I would never have been able to complete my master's studies without you all.

# DEDICATION

I sincerely dedicate this master's thesis to my parents (Rishiram Sapkota and Shova Sapkota), who always motivated and blessed me to fulfill my dreams.

# TABLE OF CONTENTS

# LIST OF TABLES

# LIST OF FIGURES

# LIST OF ABBREVIATIONS

| | |
|---|---|
| AGL | Above Ground Level |
| AOI | Area of Interest |
| CNN | Convolution Neural Network |
| CRF | Conditional Random Field |
| DGPS | Differentially correlated Global Positioning System |
| DL | Deep Learning |
| DSM | Digital Surface Model |
| DSS | Decision Support System |
| FN | False Negative |
| FP | False Positive |
| GDAL | Geospatial Data Abstraction Library |
| GIS | Geographic Information System |
| GNSS | Global Navigation Satellite System |
| GPS | Global Positioning System |
| GWM | General Weed Management |
| HR | Herbicide Resistant |
| IDE | Integrated Development Environment |
| IWM | Integrated Weed Management |
| ND | North Dakota |
| NDVI | Normalized Difference Vegetation Index |
| NIR | Near Infra-Red |
| NSRS | National Spatial Reference System |
| OBIA | Object-Based Image Analysis |
| OPUS | Online Positioning User Service |
| PCA | Principle Component Analysis |
| PIL | Python Imaging Library |
| PIP | Pixel Intensity Projection |
| PPK | Post-Processing Kinematic |
| R-FCN | Region-based Fully Convolution Network |
| RGB | Red, Green, Blue |
| SKN | Supervised Kohonen Network |
| SSWC | Site-Specific Weed Control |
| SSWM | Site-Specific Weed Management |
| SVM | Support Vector Machine |
| TN | True Negative |
| TP | True Positive |
| UAS | Unmanned Aerial System |
| UAV | Unmanned Aerial Vehicle |
| UTM | Universal Transverse Mercator |
| VRA | Variable Rate Application |
| WGS | World Geodetic System |
| WMI | Weed Management Implement |
| WMP | Weed Management Program |

# 1. GENERAL INTRODUCTION

Corn is one of the largest and most important grain crops worldwide. Corn is used for human food, livestock feed, and ethanol production. The United States (US) is responsible for around 36% of global corn production (Ort & Long, 2014). Corn-derived products are considered a major source of food for many people around the world. However, only around 10% of the corn produced in the US is used for human consumption. Corn is the biggest source of feed for livestock categories such as beef, cattle, dairy, poultry, and swine (Rimatzki, 2020). On top of that, about one-third of the corn grown in the US is now being used to produce biofuel (ethanol), which is being used daily in automobiles, commercial trucks, and even trains (US Department of Energy Ethanol Fuel Basics, 2019). Another emerging use of corn is in making polylactic acid, which is used for making biodegradable plastics (Rimatzki, 2020).

Corn is the third most important crop in terms of planted area in North Dakota (ND). In 2020, 3.73 million acres of land in ND were planted to corn (USDA/NASS, 2020). Corn in ND is most commonly planted in early May, with a row spacing of either 22 or 30 inches. The average corn plant population in the state ranges from 28,000-32,000 plants per acre, but lower population recommendations (16,000 to 25,000) are not uncommon. In the most drought-prone areas in the state, corn seeds are planted at a depth of 1.5 to 2 inches (CREC, 2019).

Weed competition is a major cause of corn yield loss in ND (CREC, 2019). Grass weed populations are usually high, and if not removed within the first three weeks of corn emergence, they can inhibit corn growth and reduce yield. Chemical herbicide is the most used option for weed control in corn crops in ND (CREC, 2019).

Weed management in corn usually consists of two chemical applications during the growing season. The first one is done at pre-emergence, often before planting, while the second one is usually done at post-emergence (V4-V5 corn growth stage) (Chahal et al., 2014; Parker et al., 2006). In order to optimize weed control efficacy and minimize the application costs, the use of both pre- and post-emergence herbicide treatments have been proven to be the most effective weed control strategy (Kudsk, 2002; Pannacci et al., 2007).

The herbicide application should be based on weed species present, crop rotation, soil type, herbicide-resistant corn technology available, and cost (Beckie, 2011; Green & Owen, 2011). However, at present, most of the farmers are using a blanket application of herbicides, to treat weeds that emerge early in the growing season. The blanket application sprays chemical herbicide uniformly across the field without considering the spatial distribution of the weeds, which often results in an overuse of chemical herbicides. This overuse can lead to economic losses, detriment to human health and reduced biodiversity (Peltzer et al., 2009). Some studies have suggested serious concerns of blanket application such as cost and effects on the environment (Martín et al., 2011; Ortiz-Monasterio & Lobell, 2007).

Site-specific weed control (SSWC) is an approach that suggests an accurate technique to spray herbicide according to the spatial variability of weeds (LÓPEZ-GRANADOS, 2011). One of the major challenges in SSWC is the accurate detection of weeds in the early growth stage (A. I. de Castro et al., 2018). Weeds are not evenly distributed across a given field, and most often they grow in aggregated patches with irregular patterns (Cardina et al., 1997). Some commercial solutions such as the autonomous weed robot by "ecoRobotix" (ecoRobotix, Vaud, Switzerland), "Deepfield Robotics" (Deepfield Robotics, Renningen, Germany), and "See & Spray" technology from John Deere Technology (John Deere, Illinois, United States) use ground-based machine vision and image processing techniques to spray only weeds. The See & Spray is the first commercially available advanced spraying technology from John Deere (John Deere, 2021).

However, this sprayer is more ideal to sense and spray herbicides weeds only under fallow ground, and the technology still has a question in terms of its ability to detect and spray post-emerged weeds accurately between and within crop rows (Wang et al., 2019).

## 1.1. Motivation for the Study

The application cost for post-emergence weed treatment in ND is around 15-20$ per acre (NDSU, 2021). In 2020, 3.73 million acres of land in ND were planted to corn (USDA/NASS, 2020). A previous study using a similar approach described in this document, showed savings of 16% on the acreage sprayed, which could reach up to 70% depending on the resolution of the prescription map generated (Flores, 2018). Taking into consideration an average herbicide cost of $17.5 per acre and the acreage planted to corn in ND in 2020, such SSWC approach could potentially generate savings in the order of $5.6-24.5 million for corn growers in the state.

The use of unmanned aerial systems (UASs) with high-resolution cameras can provide high-quality imagery, which can be processed and analyzed to obtain accurate weed distribution information (Huang, Deng, et al., 2018). Using this weed information, one can further create a weed prescription map and integrate the map with commercial sprayers. However, to achieve this, it is vital to detect weeds accurate during the early growing season of crops.

## 1.2. Objectives

The objective of this study were 1) to use UAS imagery to map and quantify weed infestation in a corn field; and 2) to integrate that map as a prescription map for weed control on a commercial size sprayer

## 1.3. Thesis Organization

This thesis is organized into six chapters: General Introduction, Literature Review, Materials and Methods, Results and Discussion, Conclusion, and Suggestions for Future Research.

The first chapter is a general introduction that describes corn production in North Dakota, associated problems related to weeds in corn fields, and the objectives of this research. The second chapter is a literature review regarding weeds, weed management, and site-specific weed management approaches, including the use of UASs for site-specific weed control (SSWC). The third chapter describes the materials and methods used in this study, from the UAS images collected to the herbicide application in the field. The fourth chapter is results and discussion, which interprets the accuracy of the research and describes possible factors that might have affected the research. The fifth chapter is a general conclusion that summarizes the results derived from chapters three and four. The sixth chapter provides some suggestions for future work.

.

# 2. LITERATURE REVIEW

A weed is a plant that grows where it is not desired (dos Santos Ferreira et al., 2017). Crop yield is directly affected by weeds as they compete with the crop for growth-limiting resources like sunlight, land, space, water, and nutrients (Thorp & Tian, 2004). The competition of weeds for these resources may result in adverse effects such as dwarfing in plant size, nutrient starving conditions, wilting, and dying of crop plants (Andreasen et al., 1996). In addition, weeds can serve as a host for insects and pathogens, which can further negatively impact crop production (Chauhan, 2020). The germination and growth of some weeds are more aggressive than crops, and they spread across land within a short period. Therefore, weed management is most important for maximizing crop yield and plant performance (Dangwal et al., 2010).

## 2.1. Weed Management Approaches

Weed management has been defined as "the process of reducing weed growth and/or infestations to an acceptable level" (Humburg & Colby, 1989). In agriculture, weed management has been accomplished with various techniques, from manual spraying/picking to tractor- and airplane-based - spraying (Carvalho, 2017). Until the last century, weed control was performed by humans visually assessing each field. Every place in a field was visually scouted for weeds, then, farmers would manually pull the weeds out (Bell, 2015; Bond, 1992). However, this manual method is very inefficient and labor-intensive. Nowadays, some common methods used to control weeds are mechanical, biological, and chemical weeding.

Mechanical weed control includes hand-pulling, hoeing, mowing, plowing, disking, cultivating, and digging weeds (Flessner et al., 2021). Biological weed management involves the use of living organisms, and livestock for weed control (Brock, 1988). However, these methods have not been effective, and the failure rate for past biological weed control techniques has been high. In addition, herbivores such as goats, cows, and sheep can provide some control over weeds, although it has not been effective (Lingenfelter, 2007). To achieve effective weed management over mechanical and biological weeding, chemical weeding was introduced.

### 2.1.1. Chemical Weed Management

The use of chemicals in agriculture in the US started in the 1870s and the chemical revolution in agricultural sciences began in the late 1950s. The inorganic salts such as sodium chloride, sodium chlorate, and arsenic salt, and inorganic sulfuric acid were used as the first herbicides (Wills & McWHORTER, 1985). These herbicides were applied at a very high rate (600-1000 pounds per acre) and they were found to be highly toxic and fire hazardous. In the 1940s there was the discovery and commercialization of 2,4-Dichlorophenoxyacetic (2,4-D), a synthetic herbicide which has been commonly used as a broadleaf herbicide for over last 60 years (Song, 2014). Since then, chemical synthetic herbicides have been the most widely adopted method for weed control in agriculture (Hamuda et al., 2016).

Spraying of chemical herbicides can significantly reduce the weeds on agricultural land and result in increased yield. However, many countries have reported residues of agrochemicals in the soil, air, water, and even in the human body (Alvarez et al., 2017; Alvarez & Polti, 2016). At the same time, the spread of agrochemicals in the environment has caused noticeable contamination of terrestrial ecosystems and human foods (Alavanja, 2009; Carson, 2002). Recent research conducted in 11 European countries showed that 83 percent of 317 soil samples contained chemical residues, mainly glyphosate, Dichlorodiphenyltrichloroethane (DDT), and broad-spectrum fungicides (V. Silva et al., 2019). Due to maximum chemical usage in agriculture, several threats to stream water and groundwater quality have been reported (Böhlke, 2002; Hallberg, 1987; Libra et al., 2020; Moyé et al., 2017; Nikolenko et al., 2018; Pérez-Lucas et al., 2019; N. S. Rao, 2018; Sasakova et al., 2018). Moreover, some classes of herbicides can have an adverse effect on biodiversity. Atrazine, one of the chemistries most widely used to control weeds, was found to enhance sterility in male frogs, and it has been banned from Europe since 2004, but it continues to be used in the US (Bethsass & Colangelo, 2006).

In addition, the massive use of chemical herbicides in agriculture has resulted in the evolution of many herbicide-resistant (HR) weeds in the last decades (Beckie, 2006; Powles & Yu, 2010), which does not support sustainable agriculture. Global herbicide usage grew by 39% between 2002 and 2011 (A. N. Rao et al., 2017). Currently, in Europe, the cost of herbicides accounts for 40% of the cost of all chemicals applied for agricultural purposes (Peña et al., 2013). Over 1 billion pounds of pesticides are used in the US every year (Donaldson et al., 2002), and according to the report of the National Pesticide Information Center, and one-third of the pesticides used in the USA are herbicide.

Because of these serious problems caused by agrochemicals regarding human health, environment, and biodiversity, there is a growing public concern over excessive use of chemicals in weed control and crop production. Finding an alternative to extensive use of chemical herbicide is a necessity for sustainable agriculture. The use of SSWC is an approach to decrease the amount of herbicide applied to crops.

**2.2. Site-Specific Weed Control Approaches**

The main idea of SSWC is to spray herbicide only to those areas where a weed is present (Wang et al., 2019). Machinery or equipment embedded with technologies that detect weeds growing near crops and act to control the weeds by considering predefined factors such as economics and effectiveness, is known as SSWC (Christensen et al., 2009). The process of SSWC is mostly based on knowledge about the spatial distribution of weeds in the field (van Groenendael, 1988).

SSWC technologies comprise three key elements: 1) a weed identifying system, 2) a weed management model, and 3) a precision weed management implementation (Christensen et al., 2009; Lopez-Granados, 2011).

### 2.2.1. Weed Identification Systems

The first important step for SSWC is weed detection. The early work on plant and weed identification systems based on remote sensing originated in the USA with NASA and military programs using satellites (Brown & Noble, 2005). Over the last 20 years, numerous weed sensing technologies have been studied and these technologies can be summarized into ground-based and aerial-based weed sensing. Ground-based weed sensing technology uses camera sensors mounted on a mobile platform to sense weeds (Wang et al., 2019). Whereas aerial-based weed sensing uses UAVs, manned aircraft, and satellites imagery to map weeds (Huang, Deng, et al., 2018).

The first step to identify weeds is image segmentation which allows for a clear distinction between plant and soil background. After image segmentation, the second step is to distinguish between crops and weeds (Christensen et al., 2009). The segmentation may be carried out by using characteristics of green foliage like the spectral reflectance in visual spectrum i.e. red, green, and blue (Woebbecke et al., 1995), spectral reflectance in the near-infrared (NIR) spectrum (Hahn & Muir, 1993), chlorophyll as a fluorescence (Keränen et al., 2003), and the non-reflectance parameters such as shape and size of leaves (Sapkota et al., 2020).

The use of multispectral and RGB sensors on a ground-based platform has shown their great ability to segment and differentiate vegetation from soil background (Thorp & Tian, 2004). Different real-time SSWC operations have been attempted by using a ground-based sensing technique that includes the use of real-time optometric, spectrometric, and RGB-NIR imaging sensors (Lin, 2010; Lopez-Granados, 2011). These techniques were successful in the ability to discriminate between vegetation and soil background; however, they could not discriminate weeds from crops during their early growth stage when the weeds and plants have similar reflectance (Lopez-Granados, 2011; Peteinatos et al., 2014). Although various studies have been able to distinguish a weed from crops using visible NIR spectroscopy, their implementation feasibility is limited to laboratory settings (Che'Ya, 2016; Dammer et al., 2013; A.-I. de Castro et al., 2012). The most effective method of weed sensing for a large agricultural area might be remote sensing using UAS or satellite. The key requirement for aerial-based remote sensing of weeds is that there should be suitable differences in spectral reflectance between weed, crop canopy, and soil background. In addition, the spatial and spectral resolution of image capturing sensors must be high enough that each pixel can differentiate between weeds, crops, and soil (Lamb & Brown, 2001). Aerial remote sensing was found to be successful to distinguish uniform and dense weed patches that have spectral characteristics larger than 1×1 meter (Brown & Noble, 2005).

Some studies have focused on the reliability and efficacy of identifying crops from weeds using simple image processing techniques such as color thresholding, filtering, and differencing (Mustafa et al., 2007; Weis, 2010; Weis & Gerhards, 2008a, 2008b). Some researchers have applied logical reasoning model to characterize weeds by a set of shape descriptors such as erosion and dilation segmentation (Aravind et al., 2015; Herrera et al., 2014). In recent years, new technologies based on deep learning and computer vision have been developed to improve the accuracy and speed of weed detection. Computer vision and deep learning-based weed detection techniques have shown their potential to discriminate weeds from crops and soil backgrounds (Liu & Bruch, 2020). Automation in crop production technology and advanced sensor-intensive technology is expected to improve production and quality with less environmental impact (Lee et al., 2010).

The use of texture feature extraction in weed detection on sugar beet by using a discrimination algorithm has shown to be efficient in discriminating weeds from crops (Bakhshipour et al., 2017). The use of image segmentation based on double thresholding 3D-Otsu's method was used to detect crop rows and then, weed and crop were discriminated by compressing the three-dimension vectors to one dimension, by making use of the principle component analysis (PCA) technique (Lavania & Matey, 2015). A sequential support vector machine (SVM) classification technique was able to discriminate between weeds and crops by using shape parameters from the image data. The technique has classified images into sub-groups such as monocotyledonous weeds, dicotyledonous weeds, and crop plants, and then the weed species were categorized into one subgroup. The characters from each sub-group were selected by SVM filters (Rumpf et al., 2012).

Currently, machine learning algorithms have been a great tool for detecting weeds. There are different convolution neural networks (CNNs) that have performed weed detection on different crops. Among all available CNNs, ResNet is regarded as the most recent CNN architecture which has the best results in terms of accuracy and speed for training image data, and the architecture was recognized at the ImageNet Large-Scale Visual Recognition Challenge (ILSVRC) in 2015 (Russakovsky et al., 2015). AlexNet, another CNN, was able to detect weeds on a soybean field using UAS imagery, and it classified the weeds into broadleaf and grasses to apply specific herbicide to those classes of weeds (dos Santos Ferreira et al., 2017). The work reported an accuracy of 97% in the detection of broadleaf and grass weeds without considering soil background. Following the ResNet CNN, a Region-based Fully Convolution Network (R-FCN) technique was developed to perform weed detection and, the module detected the weeds with great accuracy (Sarker & Kim, 2019). However, the proposed method is time-consuming for training the model. Deep Learning CNN (DL-CNN) models have been reported as remarkably accurate to detect weeds for post-emergence herbicides application on turfgrass (Yu et al., 2019). A self-supervised method that used prior knowledge about seeding patterns in a vegetable field has shown great potential to discriminate weeds from vegetable crops using hyperspectral imaging (Wendel & Underwood, 2016). However, the module worked only on static training data, which was manually labeled, and it did not show the same potential of weed detection in changing conditions of crops and plants in the image dataset. Partial Least Square Discriminant Analysis (PLS-DA) classification model, based on SVM classifier, was developed to discriminate creeping thistle, which is a perennial weed that caused yield loss in sugarbeet, but the accuracy was only good when the image data contained spectra data (Garcia-Ruiz et al., 2015).

Some companies have developed and commercialized machines that use optoelectronic sensors to measure the reflection intensity of vegetation and discriminate the vegetation from soil background. WeedSeeker, GreenSeeker, and WEED-it are some examples of these vegetation sensing products which are commercially available on the market today (Peteinatos et al., 2014; Tremblay et al., 2009). Greenseeker is designed to collect red and near-infra-red (NIR) spectral reflectance of the vegetation. NDVI derived using the sensor indicated that the Greenseeker could correlate the NDVI with other data like visual weed control, weed dry matter, etc. However, there was a positive correlation between the NDVI and weed leaf coverage (Merotto Jr et al., 2012). Similar research has been conducted using Weedseeker in California to detect weeds in the cotton field (Sui et al., 2008). This system has a sensor mounting unit on a tractor, and the spray nozzles are activated when weeds are detected through an electronic sensor. The authors reported about 80% herbicide savings on annual herbicide cost when an under-hood sensor-controlled application technology was used (Hummel & Stoller, 2002). One downside of that system is that it could not achieve discrimination of weeds from crops during the early growth stage. Recently, John Deere has released "See and Spray" sprayers that can treat pre-emergence weeds in a fallow ground without using a blanket application (Deere, 2021), but just like the WeedSeeker, the system is not capable of differentiating weeds from crops during the early growing season of plants.

**2.2.2. Remote Sensing and UAS Imagery to Implement SSWC**

In most crop-weed cases, weed treatment is recommended at the early growth stage of the crop. In this stage, mapping of weeds is the most challenging task because of four main reasons: 1) weeds have non-uniform distribution, which necessitates working at a single pixel size on the image (Robert, 1996), 2) crop and weeds have the same reflectance properties, 3) interference of soil background (Thorp & Tian, 2004), and 4) very small size of weeds which do not appear in the image due to insufficient camera resolution (López-Granados et al., 2016; Torres-Sánchez et al., 2013). The development of SSWC techniques partially relies on the use of remote sensing technology, for the collection and processing of spatial data from satellites or UAS (Steven & Clark, 2013). Satellites are a traditional platform to obtain spatial images. However, satellite imagery presents problems for many aspects of precision agriculture because of its limitations in providing enough spatial and temporal resolution (Herwitz et al., 2004). In order to overcome this, remote sensing using UASs has been widely applied in agricultural studies for mapping weed distribution (A. I. de Castro et al., 2012; Lee et al., 2010). UASs can fly at a low altitude, allowing one to capture very high spatial resolution images that can be processed to differentiate crops, weeds, and background (Xiang & Tian, 2011).

Few studies have reported the use of UASs in assessing weed distribution information. A study of a complete spray mechanism to treat aquatic weeds has used autonomous rotary-wing UAV (RUAV) to identify and locate weeds in an inaccessible location. The RUAV was equipped with an RGB camera sensor, various weed detection algorithms that process the image in real-time, and a spray mechanism (Göktoğan et al., 2010). Similarly, an object-based image analysis (OBIA) on UAS imagery has been used to characterize crop rows within a maize crop field for SSWC (Peña Barragán et al., 2012). Another study used OBIA of UAS imagery for weed mapping in an early-season maize field, using six bands multispectral camera, and the study detected crop rows with 86% accuracy (Peña et al., 2013). An OBIA study performed sunflower weed detection on color-infrared images captured at 40m and 50 days after sowing, when plants had 5-6 true leaves, while detecting weeds with an accuracy up to 91% (Peña et al., 2015). Although the reports above show good results regarding weed mapping using OBIA, the implementation of that is a complex task because this technique requires lots of information such as field structure, crop patterns, plant characteristics, hierarchical relationships, and the OBIA analysis algorithm combines object-based features such as spectral, position and orientation of weeds and crop plants.

The discrimination between weed and crop has been achieved by identifying crop rows after applying Hough transform on a UAS imagery that has only a vegetation skeleton. The rows identified in the image were further analyzed by simple linear iterative clustering (SLIC), which creates a spatial relationship of super pixels and their positions in the detected crop lines to detect intra-line as well as interline weeds (Bah et al., 2017). The mapping of weeds on a multispectral UAS image has been achieved by using supervised Kohonen network (SKN), Counter-propagation Artificial Neural Network (CP-ANN), and XY-Fusion Network (XY-F) with 98.64%, 98.87%, and 98.64% accuracies respectively. The approach used around 0.7 million features to feed the neural network models (Pantazi et al., 2017). However, these techniques require a lot of effort such as vegetation detection, classification of detected vegetation, and image labeling, which are very time-consuming tasks.

A weed monitoring system using the Hough Transform on UAS imagery has successfully detected crop rows within a sunflower field, using an SVM classifier. The research was conducted on two different flights with different altitudes and camera sensors. The three-bands and-six bands multispectral sensor were used to collect images from 30 and 100 meters, respectively. However, the result of crop row identification on a 100m image was much worse than the ones at 30m (Hervás Martínez et al., 2015). The application of Hough Transform on an orthomosaic binary image has been applied to other research studies for performing crop row detection (Bah et al., 2017; Hervás Martínez et al., 2015; H. Huang, Lan, et al., 2018; Lottes et al., 2017). Feature learning-based approaches have been used to generate image filters that allow for the extraction of features to discriminate weeds and background, on a high-resolution digital camera mounted on a UAV. The result showed the highest accuracy of the image classifier as 90% for serrated tussock. However, a lower accuracy of 70% for the tropical soda apple dataset was calculated (Hung et al., 2014).

A comparison of different machine learning models such as unsupervised, supervised, and semi-supervised techniques, have shown that the semi-supervised machine learning technique proposed for weed mapping in sunflower plants has excellent performance when using few labeled image data complemented with unlabeled data (Pérez-Ortiz et al., 2015). However, this technique is time-consuming, and it requires very high computational power.

Research for accurate weed map generation in rice fields has used an approach of semantic labeling. A fully deep convolution network, called ImageNet, has been designed and fetched with convolution filters to discriminate rice and weed on the imagery. The use of a fully connected conditional random field (CRF) has been applied for refining spatial features, with an overall accuracy of 94.45% (H. Huang, Lan, et al., 2018). However, the increased complexity of CNN leads to a decreased inference speed and executing the CRF process in the post-processing of UAV images requires a lot of time.

In order to fulfill the research gap that still exists regarding the complexity and time-consumption of weed detection algorithms in corn plants, this research used UAS imagery to map the weed infestation in a corn field. In addition, most of the existing studies are limited to the weed detection part. The authors were not able to find any published papers that encompass the full cycle to implement SSWC, from UASs image collection to spraying the field using a commercial size sprayer, such as the one described in this document.

**2.2.3. Precision Weed Management Implement (WMI)**

After weeds are sensed and the model for SSWC is ready, the WMI is the final step for SSWC. At its simplest, WMI consists of an automated tractor or sprayer that follows the SSWC model and controls the sprayer system (Christensen et al., 2009). SSWC implementation has the potential to reduce the amount of herbicide usage by 40-60%, decrease fuel consumption, and increase crop production (Jensen et al., 2012). Early research done on WMI examined the need for spatially variable herbicide application and surveyed technical literature related to geographic information systems (GIS), remote sensing, and sprayer technology (Brown et al., 1990). Development of a variable rate anhydrous ammonia application system was done in the early 1990s, by using a microcomputer and flow sensor. This technology applied the techniques of passing anhydrous ammonia through flow rate indicator and control valve in the gaseous phase (Robert et al., 1991). A prototype of a direct nozzle injection spray boom was developed with 10 system nozzles on a conventional spray boom. This model works on a variable control signal provided by small metering orifices by changing differential pressure across the orifices (Miller & Smith, 1992).

Most of the available sprayers today can operate by selective control of sections of the sprayer boom spatially, and the chemical herbicide dosages are regulated by a hydraulic pressure system. Currently, one of the most adopted technologies of SSWC implement is variable rate application (VRA). The two most common methods of implementing SSWC for VRA are sensor-based and map-based methods. The sensor-based VRA utilizes sensors to measure desired properties related to soil background or vegetation, whereas the map-based VRA uses data on spatial variation of certain parameters of soil background or vegetation, in order to optimize crop management practices (Ess et al., 2001).

Recently, map-based VRA has been widely studied and research is occurring to make this technology scalable. The three major components of map-based VRA applications are: spatial data; a model that translates the spatial data into a map; and an implementation that applies the map task site-specifically (Ess et al., 2001; Kempenaar et al., 2017). Most of the map-based VRA uses a differentially corrected global positioning system (DGPS) (Coyne et al., 2003). The maps for operating map-based VRAs need to be accurate. Remote sensing sensors, GIS programs, and some software can work together to develop maps for treating specific plants/weeds. An early study on this area showed that approximately 50% of the field only needed to be treated, and the development of a GIS-based prescription map was done based on the estimation of areas that needed for treatment or not (Lamastus-Stanford & Shaw, 2004).

**2.2.4. Economics of SSWC**

Weed control tools, like SSWC, where chemicals are applied only to areas where weeds are present, could generate some savings in inputs. Some studies in the past have demonstrated a reduction in chemical herbicide inputs by as much as 50-60% when the SSWC approach was used (Andújar et al., 2013a; Timmermann et al., 2003). An SSWC project done in the early and mid-1990s for controlling wild oat in spring wheat reported savings of $16.95 ha$^{-1}$ in controlling annual weeds when compared to the fallow land weed management method (Koger, 2001; Maxwell & Colliver, 1995). In order to determine the necessity of SSWC, research done in a soybean field indicated that only 32 hectares of land out of 60 hectares required weed treatment, but the fallow land application was made to the entire field and the treatment cost was found to be $80.03 ha$^{-1}$. A cost-saving of $36.85 ha$^{-1}$ would have been realized if only 32 hectares of land had been treated (Garegnani et al., 2000). Similar findings have been reported in SSWC of soybean fields with cost savings of $96.24 ha$^{-1}$ to $104.76 ha$^{-1}$ (Medlin & Shaw, 2000). In Europe, researchers have reported savings of €42 ha$^{-1}$ in corn, €32 ha$^{-1}$ in winter wheat, €27 ha$^{-1}$ in winter barley, and €20 ha$^{-1}$ sugar beet (Timmermann et al., 2003). Regarding weed control strategies in corn production, researchers demonstrated through simulations that SSWC uses the most profitable strategy among the seven weed control strategies simulated in the study (Andújar et al., 2013b).

# 3. MATERIALS AND METHODS

## 3.1. Study Site and Field Experiment Design

The study location was the NDSU Carrington Research Extension Center (47.51°N, 99.12°W), North Dakota, USA. The size of the research field was approximately 42 acres and was classified as a prime farmland by United States Department of Agriculture (USDA). The soil composition was Heimdall and similar soil: 42%, Emrick and similar soil: 37%, and minor components 21% (WSS, 2021). The field was planted with silage corn on May 12, 2021, with a 30-inch row spacing. Prior to planting, the field received a pre-emergence herbicide treatment (verdict) which was put down at 14 oz and 10 gallons of water on May 7. On June 19, for the post emergence treatment, we put roundup power max at 32 oz, wide match at 1.5 pt, class act 2.5 gal/100-gal water, interlock at 4 oz, and we used 15 gal of water. The experimental area was irrigated using a center pivot irrigation system.

Image datasets for this study were collected on two different dates. The image for performing SSWC during the early growing stage was collected on June 14, 2021. All the SSWC decisions were made according to these images that were collected at that time. Post-harvest imagery was collected on September 17, 2021, and it was used for evaluating the efficacy of the SSWC methodology applied to the field.

Individual images were stitched into an orthomosaic, which was the basis for analysis leading to the herbicide application in the field. Each corn row was identified using a novel Pixel Intensity Projection (PIP), a computer vision algorithm developed by the author. After the corn rows were identified, a buffer was created to cover the entire extension of the corn plants in the row, and then those corn plants were deleted from the imagery. The remaining vegetation fraction was classified as weeds. Once completed, the weed map was then converted into a weed control prescription map, which was later loaded into a commercial size sprayer to apply the treatments.

The experiment design was randomized blocks, with 6 replications of each of the following treatments: 1) the SSWC approach proposed in this study; and 2) non-SSWC currently used by farmers (blanket application across the field). Each plot measured 136.6 ft (sprayer boom width) x 400 ft (width x length).

Once the prescription map was loaded into the sprayer, the sprayer was driven over the field, and nozzles were shut off automatically, based on the prescription map created from the imagery, on the areas free of weeds. The basic block diagram of the research method is presented in Figure 1.

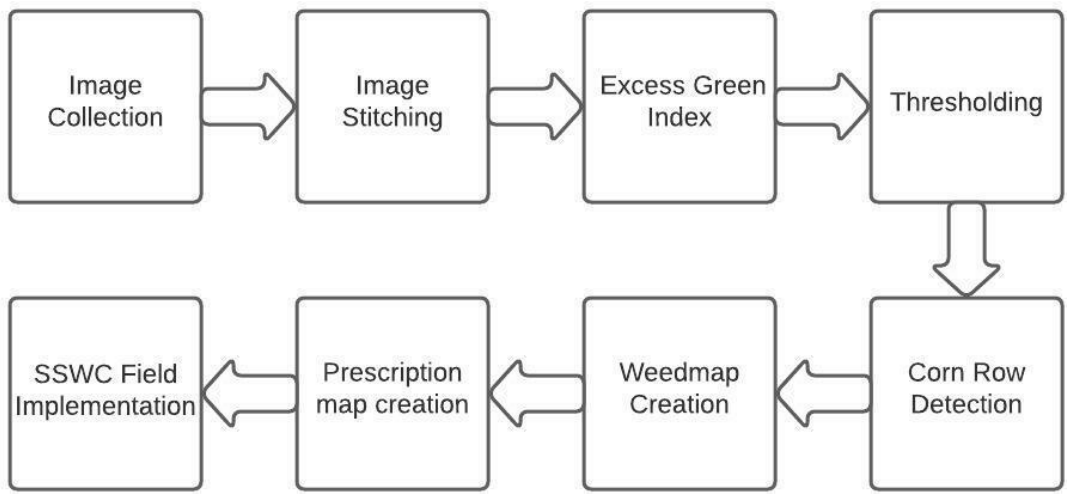

**Figure 1. Block diagram showing the process from image collection to herbicide spray application using SSWC approach**

### 3.2. UAS Image Collection and Geotagging

A DJI Matrice 600 Pro (M600) (DJI, Shenzhen, China) was outfitted with a Sony Alpha 7R II 42 Megapixel RGB camera (Sony City, Tokyo, Japan) to capture the aerial images of the field (Figure 2). The Matrice 600 Pro is a hexacopter, that includes the batteries, has a total weight of 9.5 kilograms, a maximum takeoff weight of 15.5 kg, and an advertised flight time of 35 minutes (in reality more like 20-25 minutes). That platform was chosen because of its heavy-lift capacity and its support to third-party hardware, such as gimbal and cameras.

The camera mounted on the M600 was a Sony Alpha 7R II mirrorless camera, which has a sensor size of 7952 × 5304 pixels (42.4 megapixels), and a focal length of 35mm. The integration of the camera on the drone was made by FieldofView LLC (Fargo, North Dakota, USA), which makes a device (GeoSnap) that allows one to trigger the camera and geotag the images with PPK (post-process kinematic) accuracy (2 cm).

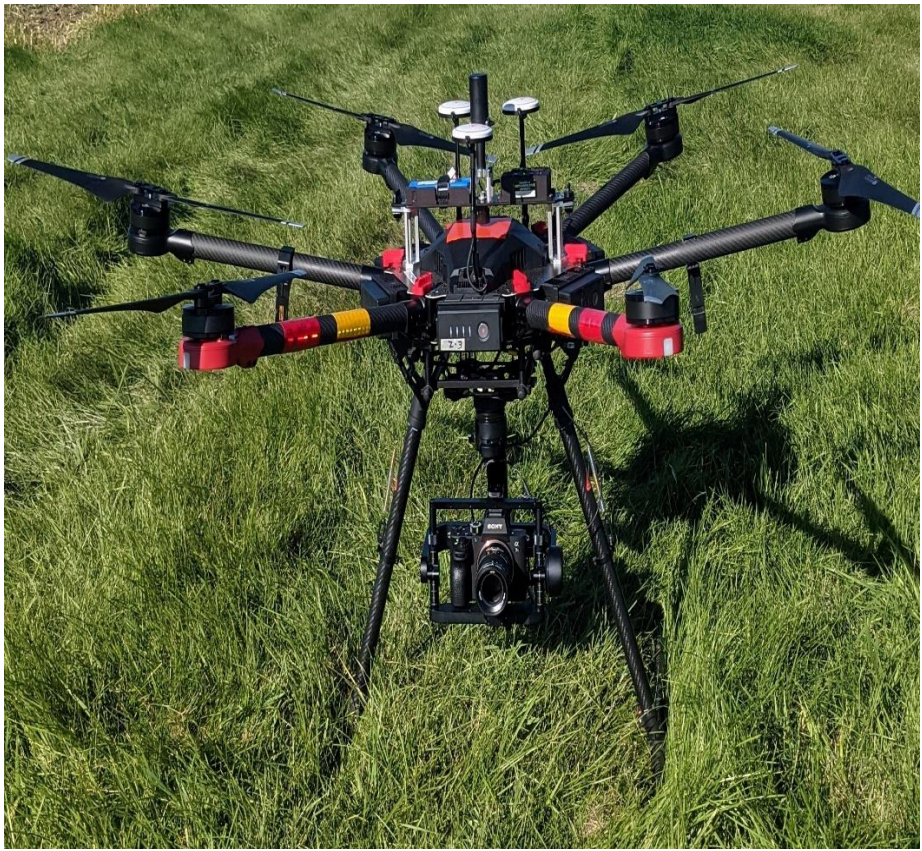

**Figure 2. DJI Matrice 600 Pro with a Sony Alpha 7R II, 42.4-megapixel RGB camera mounted on it. The drone is loaded with 6 batteries and a Geosnap PPK system is integrated to the UAS by mounting it on the top part of drone.**

The UAS was flown autonomously using the Pix4D Capture app (IOS version) (PIX4D, Prilly, Switzerland), which was used to create and fly the missions. Flights were carried out at 350 feet above ground level (AGL), with the camera at nadir position, with 75% overlap both front and side, and the UAS speed was adjusted (by the app) to allow 1-2 seconds interval between pictures. It took around three flights to fly the experimental field, for a total of 2251 images.

The GeoSnap PPK device was mounted on top of the M600 Pro (Figure 2) and it is shown in more detail in Figure 3. The system is connected directly to the camera and the settings (flight altitude and overlap) in that device were identical to the ones set for the mission in Pix4Dcapture so that the app would fly the drone at the correct speed and space between passes, while the GeoSnap system would trigger the camera to the correct locations. The system logs raw GNSS data for each image, which allows one to generate centimeter-accurate post-flights geotags using PPK software and data from a base station. Due to the distance between the research field and the nearest CORS (Continuously Operating Reference Stations; Cooperstown, ND), which would cause a degradation of the geotag accuracy, an iG4 NSS (iGage Mapping Corporation, Salt Lake City, USA) base station was used at the edge of the field to serve as a reference during the workflow (not described here) to implement the PPK corrections to the geotags. The base station tracks 432 channels of GPS, GLONASS, BeiDou, and Galileo satellite constellations. As per instructions of the manufacturer, the base station was set up a minimum of two hours before each flight. That base station runs approximately 6.5 hours on a 3,400 mAH battery (*IG4 GNSS*, 2021).

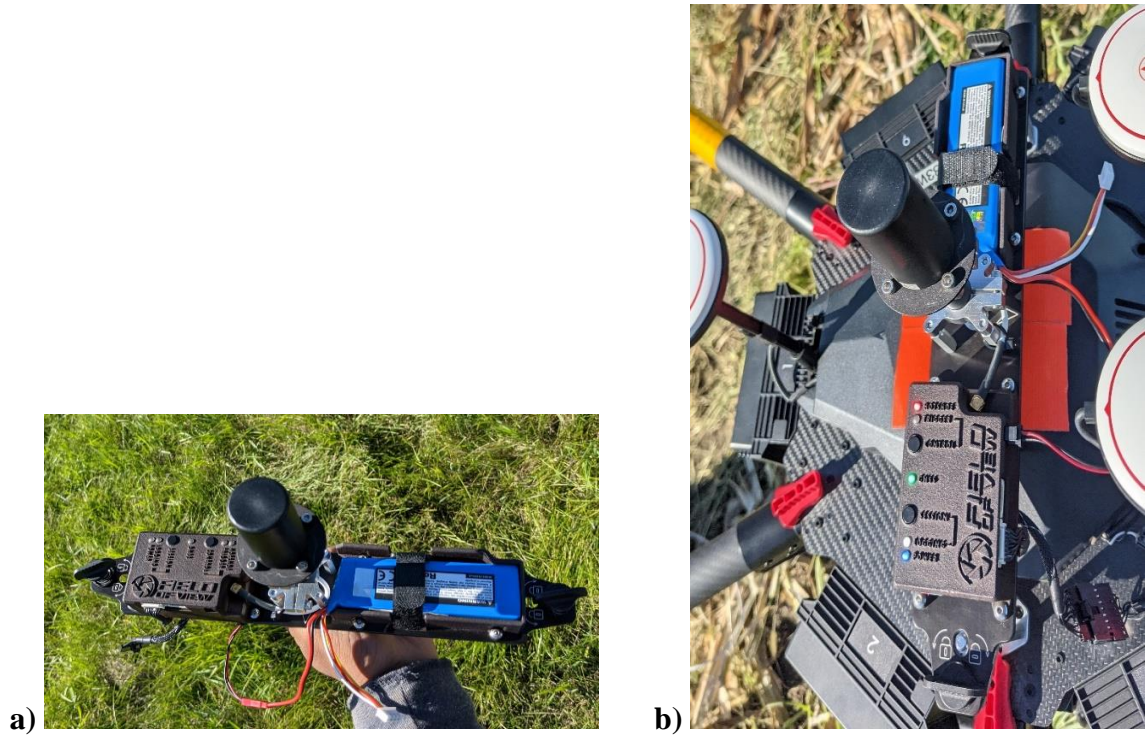

**Figure 3. a) Field of View Geosnap PPK before mounting to the drone, b) Field of View Geosnap PPK mounted on top of drone for triggering aerial geotags at each image capture.**

## 3.3. Image Pre-Processing and Stitching

EzSurv software (Effigis, Montreal, Canada) was used to generate geotag information with an accuracy of 2-cm for latitude and longitude, and twice that for elevation, for the images prior to the stitching process. That information was then saved into a .csv file which was specially formatted according to the requirement of an image stitching software, Pix4D (PIX4D, Prilly, Switzerland). Using the PIX4Dmapper educational license, the images were processed into an orthomosaic, which was georeferenced according to the information provided through the .csv file generated by the EzSurv software.

The processing was done in a desktop computer with Windows 10 Pro, version 20H2, a 64-bit operating system with 132 gigabytes of RAM. The desktop was provided with two switchable GPUs. One was Intel(R) UHD Graphics 630, and the other was NVIDIA GeForce RTX 2080 SUPER. The whole process of image stitching in Pix4Dmapper was completed in 6 hours and 25 minutes.

Figure 4 shows the orthomosaic output from Pix4Dmapper and the experimental areas where the treatments were applied. The ground sampling distance of the orthomosaic was 0.63 cm. This orthomosaic served as the basis for all the subsequent analyses to implement the SSWC approach described in this document.

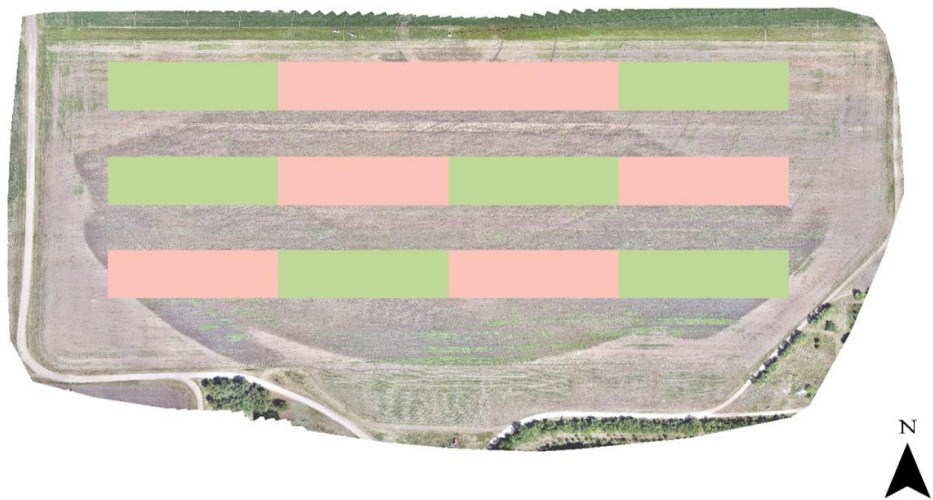

**Figure 4. Orthomosaic generated from a corn field after images were stitched in Pix4Dmapper. Images were captured by a Sony Alpha 7R II, flown at 350 ft AGL. The colored box shows the placement of the SSWC (green) and no-SSWC (pink salmon) treatment plots.**

### 3.4. Vegetation Identification

The orthomosaic generated from Pix4Dmapper was further processed in ArcGIS Pro (Esri, California, USA). The first step in the process was to segment the vegetation fraction on the imagery from the remaining background. A common vegetation index used in the literature to accomplish that kind of segmentation is the excess green index (ExGI, Equation 1) (Torres-Sánchez, Pena, et al., 2014). The ExGI highlights the green portion of the spectrum allowing one to segment the vegetation from the remaining imagery background by using a threshold value. The vegetation region of interest was then converted to a binary image using the reclassify tool in ArcGIS Pro.

The expression for ExGI calculation is:

$$\text{ExGI} = 2g - r - b \qquad \text{(Equation 1)}$$

where,

$$r = \frac{R}{R + G + B}$$

$$g = \frac{G}{R+G+B}$$

$$b = \frac{B}{R+G+B}$$

where r, g, and b are the normalized values of the bands red, green, and blue respectively; and R, G, and B represent the digital number values of the red, green, and blue color band from the orthomosaic.

To segment the green vegetation from the background, an exploratory investigation was done in ArcGIS Pro to identify the ExGI threshold that would distinguish those two classes. The threshold ExGI of 0.08 was found to be appropriate for the segmentation. Using the "Raster Calculator" tool, a binary image (ExGI <= 0.08 = 0, and ExGI > 0.08 = 1) was created from the ExGI raster. Since the vegetation class (ExGI > 0.08 = 1) was one of interest, the background values, or zeros, on the binary imagery were deleted.

**3.5. Algorithm Development for Corn Row Detection**

To expedite the processing time, in this study only the portions of a binary image that were within the boundaries of the plots assigned to the SSWC treatment were processed for crop row identification. There was no need to perform that task on the other parts of the imagery since they were going to receive a blanket application of herbicide.

The workflow to detect the corn rows was implemented using Python libraries (Figure 5). The main libraries used for the workflow were OS, NumPy, PIL (Python Imaging Library), Rasterio, GDAL (Geospatial Data Abstraction Library), Matplotlib, OpenCV, and SciPy.

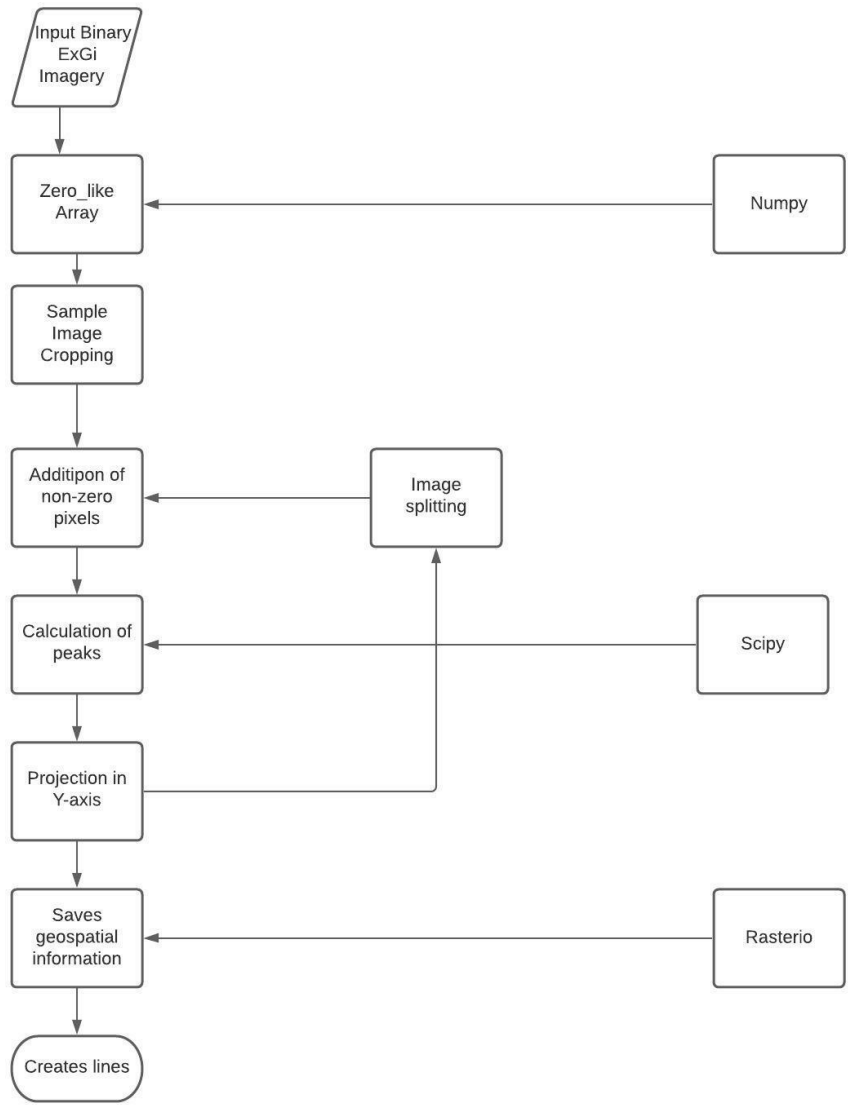

**Figure 5. Workflow diagram for corn row identification from UAS imagery using Python libraries and computer vision techniques**

The main goal of this operation was to generate an output image that would contain only lines representing the corn rows. The first step to proceed for line detection was to prepare a zero-like array using NumPy, which returned a new array of given shape and type, filled with zeros.

Once the zero-like array was created, a function was defined as "cropped_image" which cropped the input imagery into smaller size (3000 X 2000 pixel). This function was set with a combination of NumPy-SciPy libraries. The NumPy variable "sum_y" sums up all non-zero-pixel values into one value for each corn row. The SciPy variable "peaks" calculated the local maximum values (peaks) in the array of Y-axis summed values. Figure 6a shows one of the cropped regions of 3000 X 2000 pixels and Figure 6b shows the calculated local maximums (peaks).

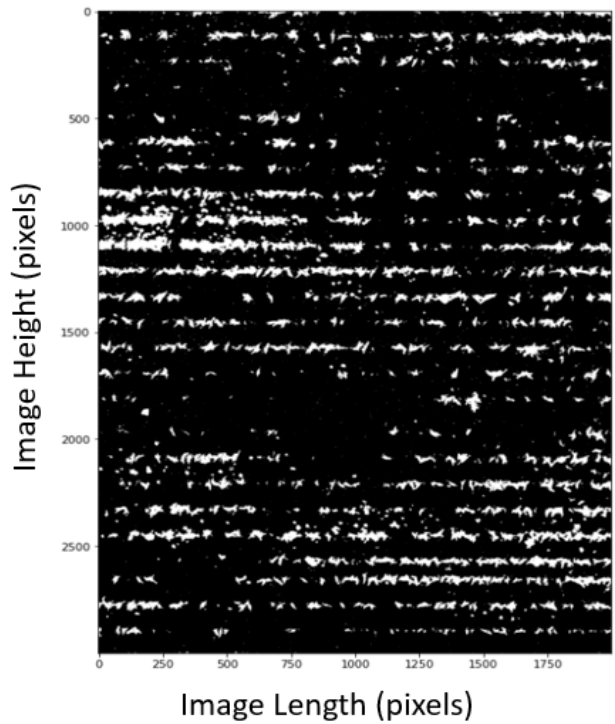

a)

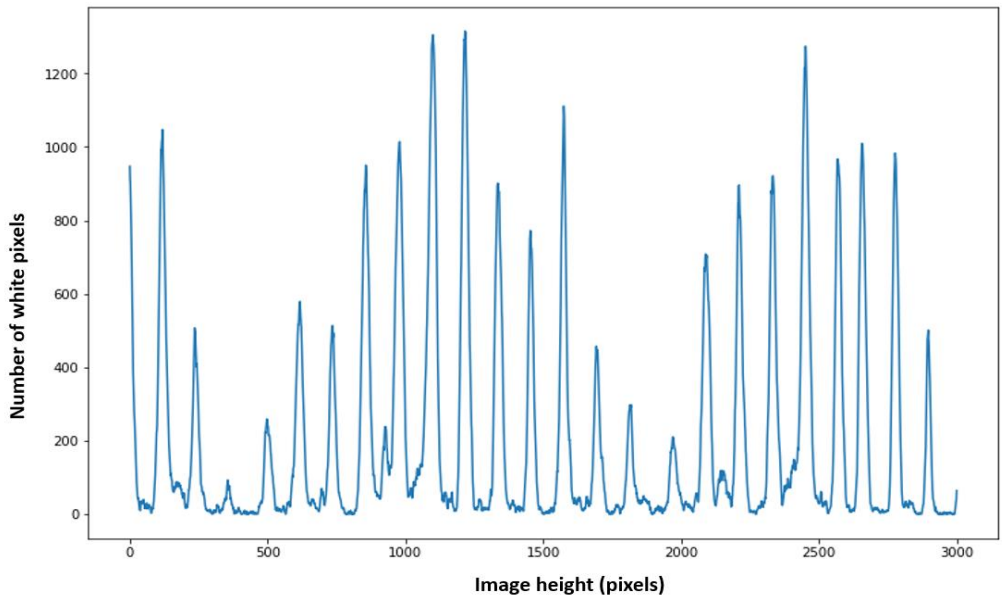

b)

**Figure 6. a) A section of the binary ExG imagery (3000 X 2000 pixels) with the horizontal orientation of rows, and b) local maximums (or peaks) for each corn row on the section of imagery. The imagery was collected by a Sony Alpha 7R II, flown at 350 ft above ground level**

In addition, a "for loop" was created for calculating the local maximums throughout the input binary imagery by splitting the whole image into small samples of 3000 X 2000 pixels. This splitting was set to start from the top left corner and end at the bottom right corner of the input imagery. The position of each small sample image was counted by using a function inside the "for loop" in accordance with the X-axis. The X-axis position was indicated by a variable "j" and the position of these samples in the Y-axis was indicated by a variable "k". The area of the whole input binary imagery was stored in a variable called "field_image", and the small cropped images were stored as "cropped_image". All the area of cropped_images within the area of field_image was selected. Additional zero arrays were created, which had the same size as the small cropped images. These zero arrays were created because the output of interest was just the lines of corn rows and nothing else. The local maximums were calculated for all small images that were split earlier. Then finally, lines were created in each local maximum. The width of the line was set to range between -3 and 3 pixels in each calculated local maximum. Then the resulting image was extracted with its associated geospatial information, using the Rasterio library.

The precision and accuracy of the corn row detection method were calculated by using the standard formula given as:

$$\text{Precision} = \frac{\text{True Positive(TP)}}{\text{True Positive(TP)} + \text{False Positive(FP)}}$$

$$\text{Accuracy} = \frac{\text{True Positive(TP)} + \text{True Negative(TN)}}{\text{True Positive(TP)} + \text{False Negative(FN)} + \text{True Negative(TN)} + \text{False Positive(FP)}}$$

where the algorithm depicted four possible outcomes as True Positives (TP), True Negatives (TN), False Positives (FP), and False Negatives (FN). These outcomes were obtained from the visual comparison of the algorithm's output imagery with the original RGB orthomosaic. The specification of each outcome was as follows.

TP = When the actual value is positive, and the algorithm also predicts a positive value (Google, 2021). In our case, TPs are the total number of predicted lines that exactly laid over the actual corn rows.

TN = When the actual value is negative, and the algorithm predicts a negative value (Google, 2021). In our case, the background in the imagery was negative. Since the algorithm did not detect any background as a background value, TN is zero in our study.

FP = When the actual value is negative, but the algorithm predicts a positive value (Google, 2021). There were some lines generated where there were not any corn rows, and the total number of those kinds of lines were considered as FPs.

FN= When the actual value is positive, but the algorithm predicts a negative value. There were some places where actual corn rows were present, but the algorithm did not generate line over that row. These kinds of cases were considered FNs.

**3.6. Mapping the Weeds Across the Treatment Plots**

In order to create the weed prescription map, the next step was to identify the vegetation fraction that would be classified as weeds. This part of the analysis was carried out in ArcGIS Pro. Our approach was to identify the corn plants first and delete those from the imagery so that the remaining vegetation in the field was considered weeds. To implement that approach, buffers of 3.5 inches were created on both sides of the rows in SSWC treatment plots. The buffer size was determined after a few tries, and the 3.5 inches size on both sides shows that it was wide enough to cover the vast majority of corn plants in the rows, without covering weeds between the corn rows. Figure 7a shows a portion of the SSWC test plot where corn row lines were buffered. Once the buffering regions were in place, all the vegetation under the buffer area was deleted, creating a data layer with only weeds (Figure 7b).

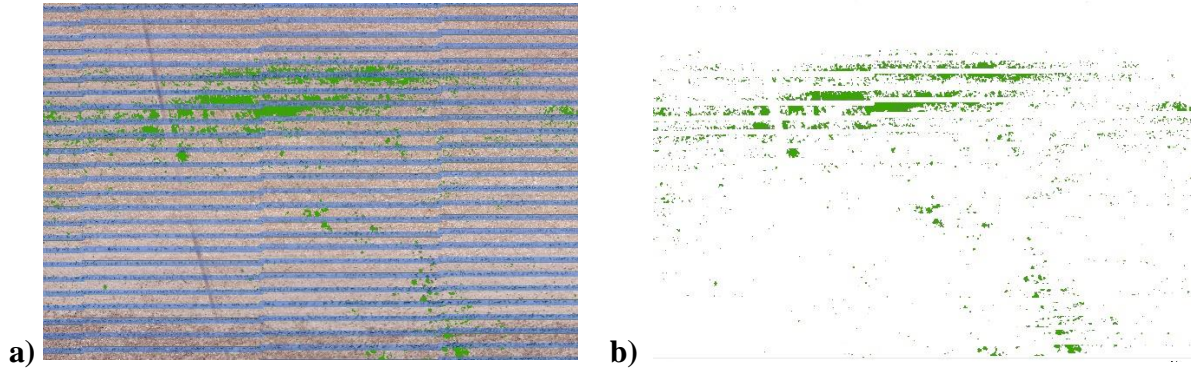

**Figure 7. Buffered (3.5-inch on both sides) corn rows (a) that were removed from the imagery to create the weed map across the field (b). Imagery collected by a Sony Alpha 7R II, flown at 350 ft above ground level**

To create the weed control prescription map, first a grid of cells of 10 ×10 ft was overlaid on the imagery. The grid was created in ArcGIS Pro and saved as a shapefile. The idea behind using these grid cells was to provide a buffer area for the spraying (if there is at least one weed that is completely within the cell, the whole cell would be sprayed) and to control weeds within the row (each grid cell would cover, in theory, four corn rows). Cells that would be sprayed were then assigned an application rate of 15-gal ac$^{-1}$, while ones free of weeds had a rate of 0-gal ac$^{-1}$ assigned to them. Since the weed pressure in the field was unusually high in the 2021 growing season, using a 10 x 10 ft grid size would result in spraying almost 100% of the cells, which would not allow for enough replications of a non-spray comparison. For that reason, and due to the individual nozzle control feature on the sprayer, the research team opted to use a cell size of 1.67 (distance between nozzles) x 10 ft. That resulted in an increased number of grid cells free of weeds, resulting in an average value of 35% of the cells not being sprayed across the SSWC treatment.

To create the prescription map, the cells from the shapefile created by the process above with a "0-gal ac$^{-1}$" rate for all six SSWC plots were selected and exported as a new shapefile. That shapefile was then overlapped with a shapefile containing the plots (12 total) for both treatments, where the field "application_rate" had been previously populated to 15-gal ac$^{-1}$. The overlapping operations highlighted the cells on the second shapefile (with all 12 plots) where the nozzles should be turned off (0-gal ac$^{-1}$), so the "application_rate" was recalculated to reflect that value. Figure 8 shows the shapefile, which would be converted on a prescription in a later step, resulting from that process (green cells = 15-gal ac$^{-1}$ and red cells= 0-gal ac$^{-1}$).

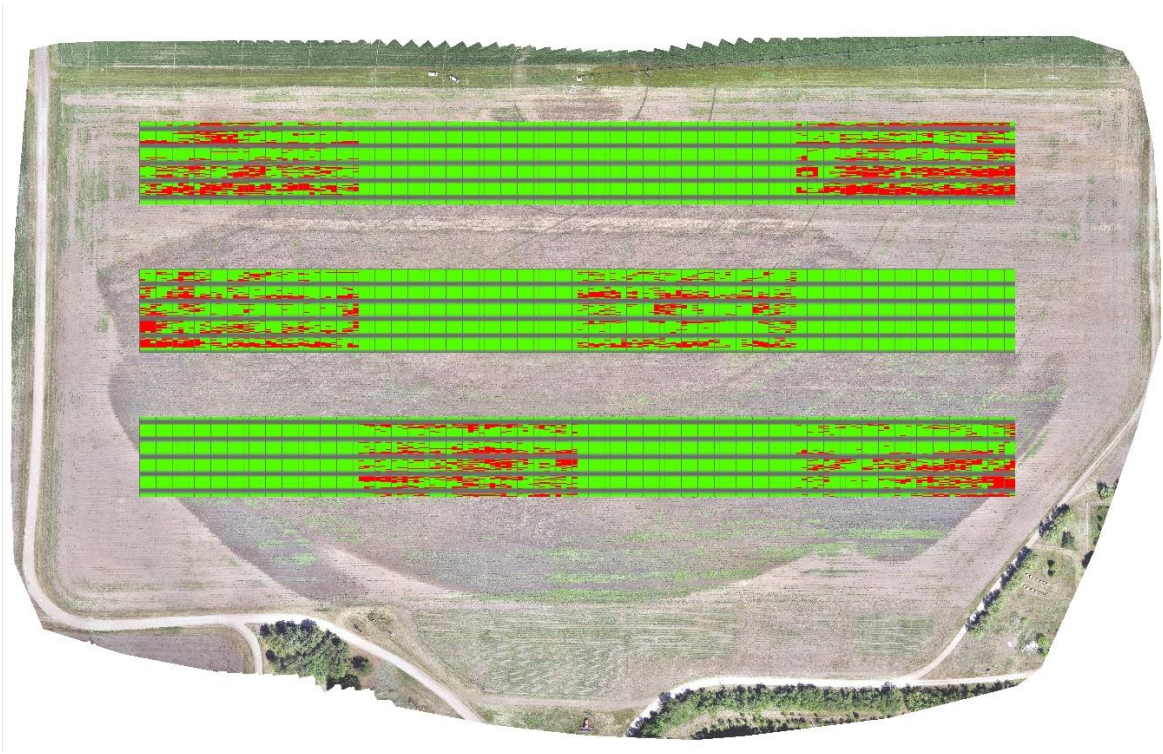

**Figure 8. Variable rate prescription map (green cells = 15-gal ac$^{-1}$ and red cells= 0-gal ac$^{-1}$) generated in ArcGIS Pro based on UAS imagery collected over a corn field by a Sony Alpha 7R II, flown at 350 ft above ground level**

### 3.7. Integration of Weed Map into a Commercial Sprayer

A Case IH Patriot 4440 self-propelled sprayer (Racine, Wisconsin, USA), model year 2021, was used in this study. Among other specifications, the sprayer was provided with an AIM command FLEX system, which is the latest technology from case IH that enables consistent, flexible, and accurate application, regardless of speed and terrain. The AIM Command FLEX uses pulse width modulation (PWM) technology and enhances control of liquid product flow and pressure more accurately than conventional rate controllers. The technology delivers the correct rate and droplet size across the field by minimizing skips and overlaps. In addition, the technology enables instant on/off of individual nozzles, with a nozzle valve diagnostic system (Case IH, 2021).

The sprayer was set up with a Raven RX1 RTK receiver ((Raven Technology, Sioux Falls USA), which operates at a frequency of 10Hz. Following Case IH engineer's advice, the application speed for this study was kept below 7 mph (6.5 mph actual speed application) to allow for the sprayer to maintain position accuracy resolution of at least one foot per 1/10$^{th}$ of a second or 10 ft per second. That cab computer requires a certain folder structure, so it can read prescription maps (Rx) from an external storage device, such as a thumb drive. AgSMS Advanced (Ag Leader, Iowa, USA) software was used to create that folder structure. The reason to use AgSMS Advanced was that it can interface with a variety of cab computers used on precision agriculture equipment. Once the Rx map was brought into the Viper 4+ display, anything outside the area of the Rx map was set to be sprayed with a 15- gal ac$^{-1}$ rate, which is the same rate that one would use for a blanket chemical application. Figure 9 shows the Patriot 4440 sprayer used in this research.

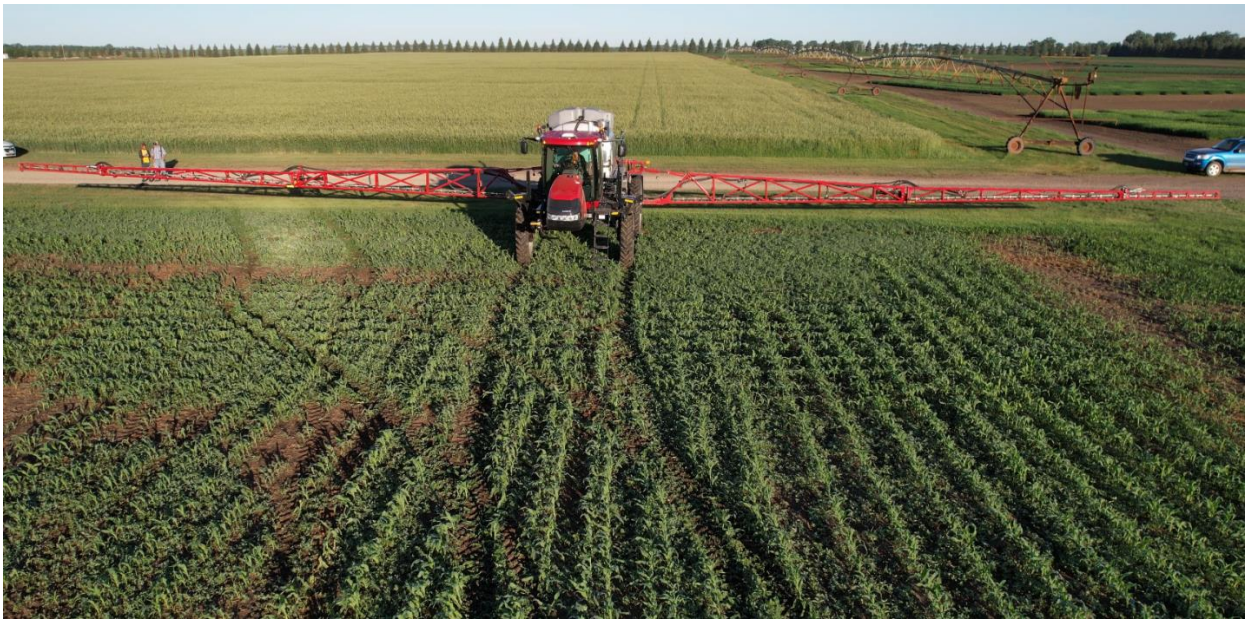

**Figure 9. Case IH Patriot 4440 series sprayer used to implement site-specific weed control in a corn field. The sprayer was equipped with an AIM command FLEX system, RTK GPS receiver, Viper 4+ cab computer, and 136.6 ft wide boom**

### 3.8. Assessing the Spraying Performance

Once the spraying was completed, the "as-applied map" was downloaded from the cab computer. In order to evaluate the accuracy of the as-applied map, that map was compared with the Rx map provided to the sprayer. The initial idea was to use those layers to assess the accuracy of the application around (inside and outside) of the polygon delineated by the Rx map. After an initial visual assessment, it was decided that such an approach would not work since the spraying pattern around the no-spray polygons was not consistent across the field. To overcome that issue, sprayer performance was assessed based on the spatial area that was sprayed and not sprayed between the Rx maps and the as-applied map. The total area which was not-sprayed in the as-applied map was considered as Measured Value, and the total area which was set to be not-sprayed in the prescription map was considered as Expected Value. The accuracy of application in terms of spraying was calculated using equation 2.

$$\text{Accuracy} = \frac{\text{Measured Value}}{\text{Expected Value}} \times 100\% \qquad \text{(Equation 2)}$$

In order to evaluate the overall effectiveness of this study, post-harvest imagery was collected on September 21, 2021. The process of image collection, image stitching, georeferencing, and ExGI calculation was done by following the similar steps that were applied for early growing season imagery. The image collection for post-harvest was done using the same UAS (DJI M600 Pro) and camera (Sony alpha 7R II) flown at 350 feet AGL. The whole process of image stitching in Pix4Dmapper for post-harvest imagery was completed in 8 hours and 35 minutes.

The main goal of collecting post-harvest imagery was to verify the effectiveness of the SSWC approach proposed in this study, by measuring the weed coverage area (ground coverage of green plants) in the cells that were not sprayed and comparing that with the cells that were sprayed `in the summer. To be considered an efficient approach to control weeds, the ground coverage by weeds in the no-sprayed cells should be less or equal to the ones that were sprayed in the spring. If the no-sprayed cells in the spring consistently result in more weeds in the fall, farmers might not be willing to use the proposed SSWC because it will increase weed seed banks in the soil.

One of the issues found with the post-harvest imagery was that it did not overlap as expected on top of the imagery collected in the spring, even though both were collected with PPK accuracy level (2 cm). In order to identify the exact SSWC treatment plots, points on the ground on both image collection dates were identified. Once completed, the polygons of SSWC and NO-SSWC treatment plots were moved manually to their location by visually comparing the known points with early-season imagery using ArcGIS Pro. To further validate that the location of the plot, the distance between the top of each plot and the first corn row on the northern side of the field was measured and found to be equal (round off to 2 digits in feet) in both early season and post-harvest imagery. In addition, all the plots were compared with the known references at the East, West, and South parts of the field, and necessary adjustments were made for each plot.

In order to evaluate the amount of weed distribution in the SSWC and NO-SSWC test plots from the post-harvest imagery, the ExGI was calculated using the same method which was applied for early season imagery. A threshold of 0.07 was chosen (after trying other values) and applied to the imagery in turn to segment the green pixels (weeds) from the imagery.

### 3.9. Statistical Analysis

The post-harvest imagery was used to evaluate the effectiveness of the approach proposed here for SSWC. The area of weeds in six SSWC and six NO-SSWC test plots was used for the statistical analysis. The datasets were analyzed using SAS PROC MIXED (SAS Institute, Cary, USA) with a mixed procedure using REML (restricted maximum likelihood) estimation.

Altogether, the field trial had six replications for each treatment (SSWC and no-SSWC). The sum of weed area in the six SSWC treatment plots was compared with the sum of weed area in the six no-SSWC treatments. The mean separation between the weed area of SSWC and no-SSWC test plots was conducted using a pair-wise T-test with a significance level of 0.05 (p-value= 0.05).

# 4. RESULTS AND DISCUSSION

## 4.1. Corn Row Detection Accuracy

The product of the Pixel Intensity Projection algorithm (proposed by the author) is a binary image, where lines represent corn rows. Figure 10 shows all the corn rows generated by the algorithm for the SSWC treatments plots overlaid on top of the orthomosaic for the corn field. Due to the size of each plot (136.6ft × 400ft) and the zoom level in Figure 10 a, one cannot see the individual corn rows in any of the SSWC plots. Figure 10 b shows a detailed view of the corn rows generated by the proposed algorithm for the plot SSWC-1 from Figure 10 a.

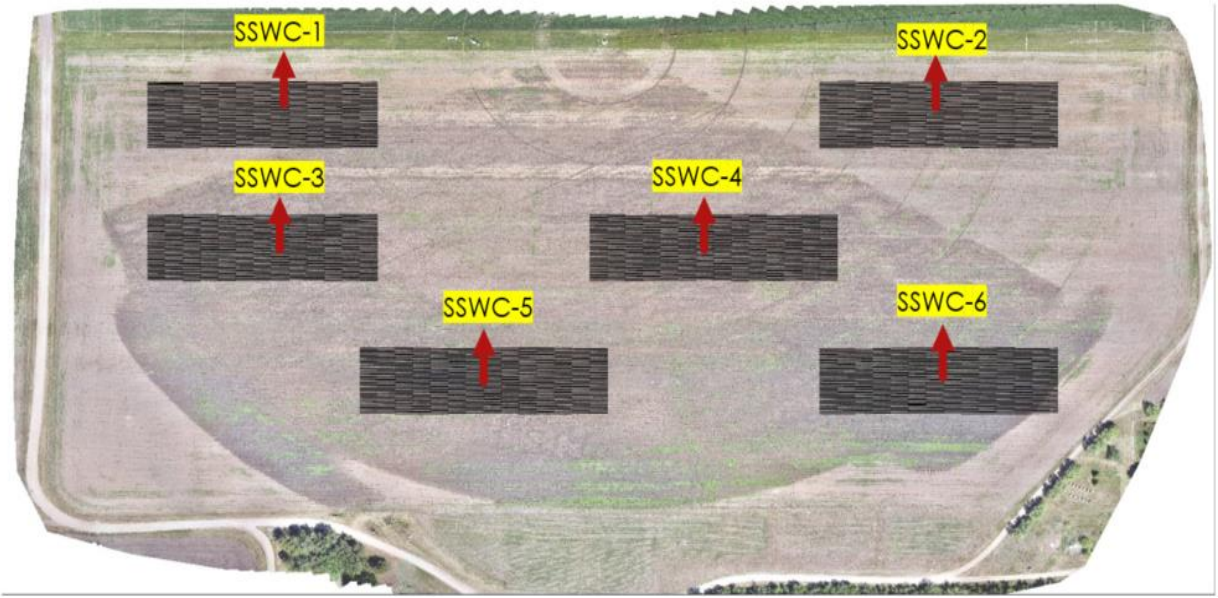

a)

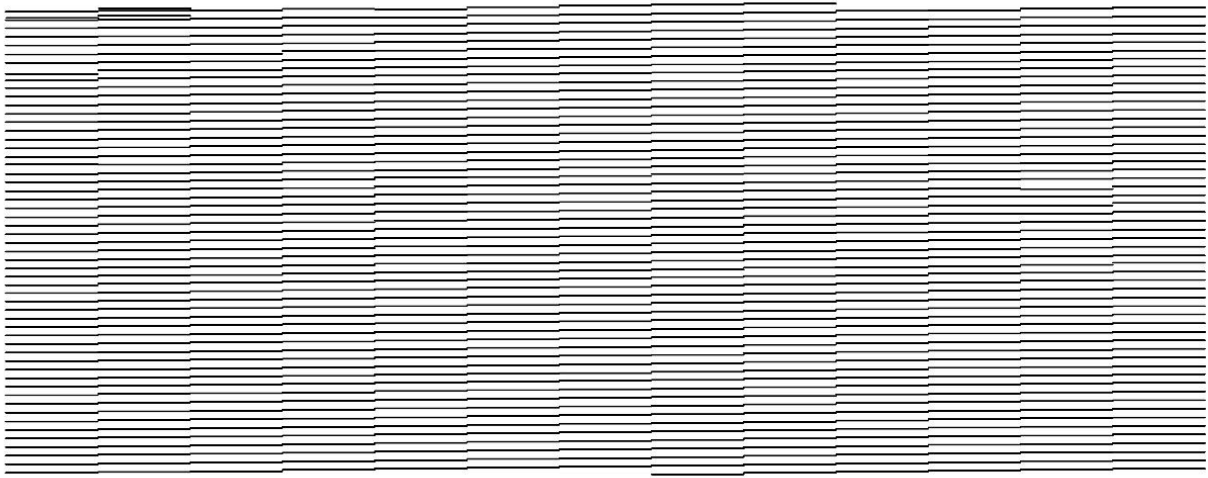

b)

**Figure 10. a) Identification of corn rows on site-specific weed control (SSWC) plots by using the Pixel Intensity Projection algorithm and ExGI imagery from a corn field. Imagery captured by a Sony Alpha 7R II camera flown at 350 ft AGL; b) Detailed view of detected corn rows lines for the test plot SSWC-1 (Figure 10 a), identified by using the Pixel Intensity Projection algorithm and ExGI imagery from a corn field. Imagery captured by a Sony Alpha 7R II camera flown at 350 ft AGL**

The lines generated from the algorithm were compared with the actual corn rows by visually assessing the orthomosaic in the ArcGIS Pro. Figure 11 shows a sample that was used to compare actual corn rows and the identified crop rows as the line.

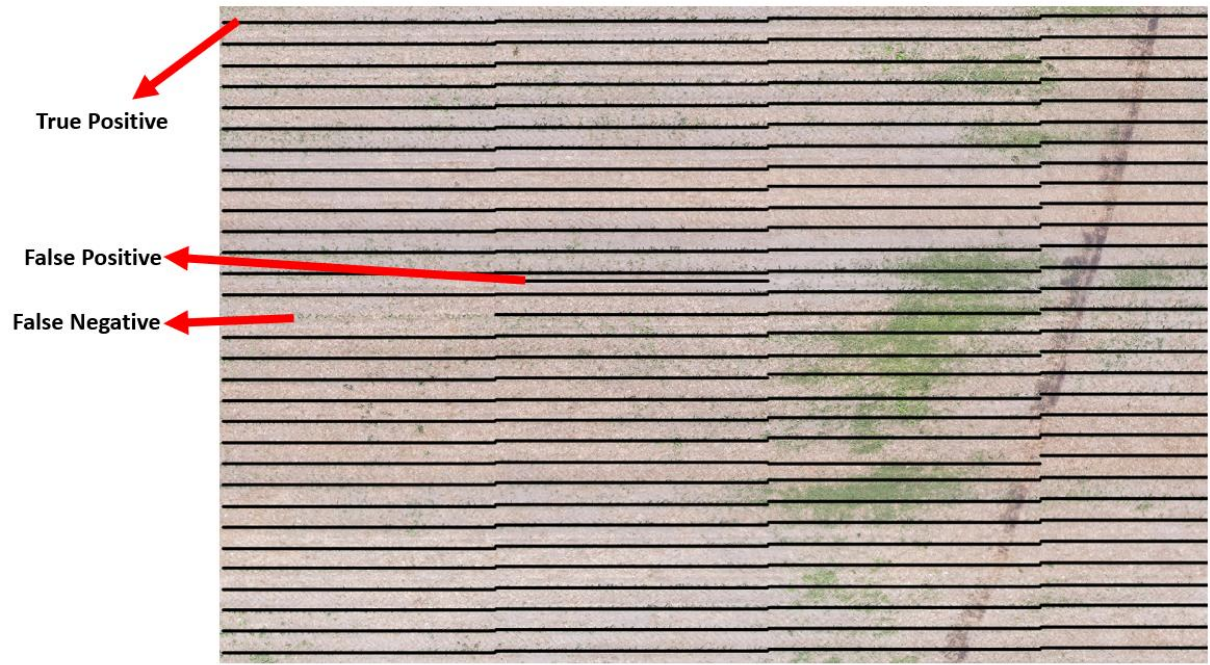

**Figure 11. Comparison of actual corn rows with the identified lines created by the Pixel Intensity Projection algorithm on a corn field imagery captured by Sony Alpha 7R II, flown at 350 ft AGL**

The number of True Positive (TP), False Positive (FP), True Negative (TN) and False Negative (FN) was manually counted for the six SSWC test plots (Table 1). Since the algorithm was not designed to detect any background, True Negative (TN) is always zero in such a case.

**Table 1. The count of TP (True Positive), TN (True Negative), FP (False Positive), and FN (False Negative) generated by the Pixel Intensity Projection (PIP) algorithm in each SSWC plot created on a corn field imagery captured by Sony Alpha 7R II, flown at 350 ft AGL.**

| Test-Plot | TP[1] | TN[2] | FP[3] | FN[4] |
|---|---|---|---|---|
| SSWC-1 | 392 | 0 | 3 | 1 |
| SSWC-2 | 382 | 0 | 1 | 3 |
| SSWC-3 | 386 | 0 | 4 | 0 |
| SSWC-4 | 385 | 0 | 2 | 1 |
| SSWC-5 | 384 | 0 | 4 | 2 |
| SSWC-6 | 384 | 0 | 1 | 1 |
| TOTAL | 2313 | 0 | 15 | 8 |

[1]TP indicates the number of lines identified by the PIP algorithm that was exactly above the actual corn rows;
[2]FP indicates the number of lines identified by the PIP algorithm that was generated where there were no corn rows;
[3]FN indicates the count where corn rows lines were not generated above the actual corn rows; and
[4]TN indicates the value of background detected by the algorithm (which is zero).

The comparison of the lines identified by the PIP algorithm with the actual corn rows showed that 2313 lines were exactly on top of the corn rows, 15 lines were generated where there were no corn rows, and 8 lines were not identified by the algorithm where there were corn rows.

The accuracy of the proposed PIP algorithm was calculated as 99.01% which shows that the algorithm was able to detect 99.01% of actual corn rows as lines. Similarly, the precision of the proposed PIP algorithm was calculated at 99.35%, which shows that the algorithm was able to predict 99.35% of lines that were relevant to original corn rows however, in some regions, double lines were generated.

Figure 12 shows a sample image where the algorithm failed to detect a corn row as a line. The corn plants on the third row on Figure 12 were much smaller in size when compared to the corn plants of other rows, which can be translated into less pixel intensity level in the ExGI imagery, which led the algorithm to fail when trying to find the local maximum or peaks for that row.

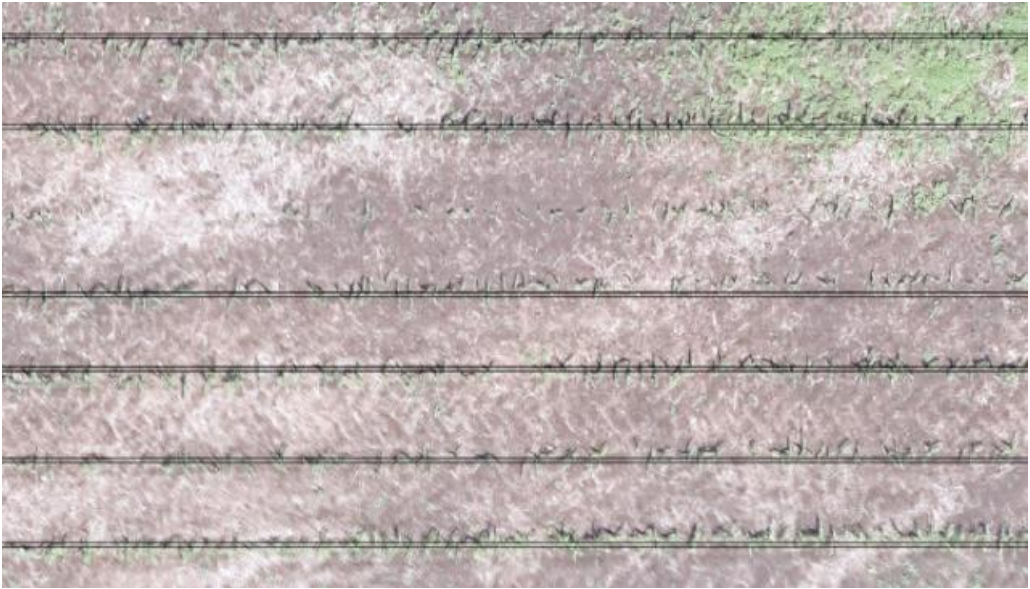

**Figure 12. Sample image where the proposed PIP algorithm failed to detect an existent corn row as a line on the ExGI imagery of a corn. Imagery captured by Sony Alpha 7R II, flown at 350 ft AGL.**

In some regions, the algorithm identified double lines over a single corn row (Figure 13). The reason behind this error is that the actual corn rows were not perfectly straight. As shown on Figure 13, the third corn row is slowly deviating downward toward the left side of the image, which indicates that the row was not straight. The algorithm calculated two local peaks for that row and projected the lines over the two calculated peaks.

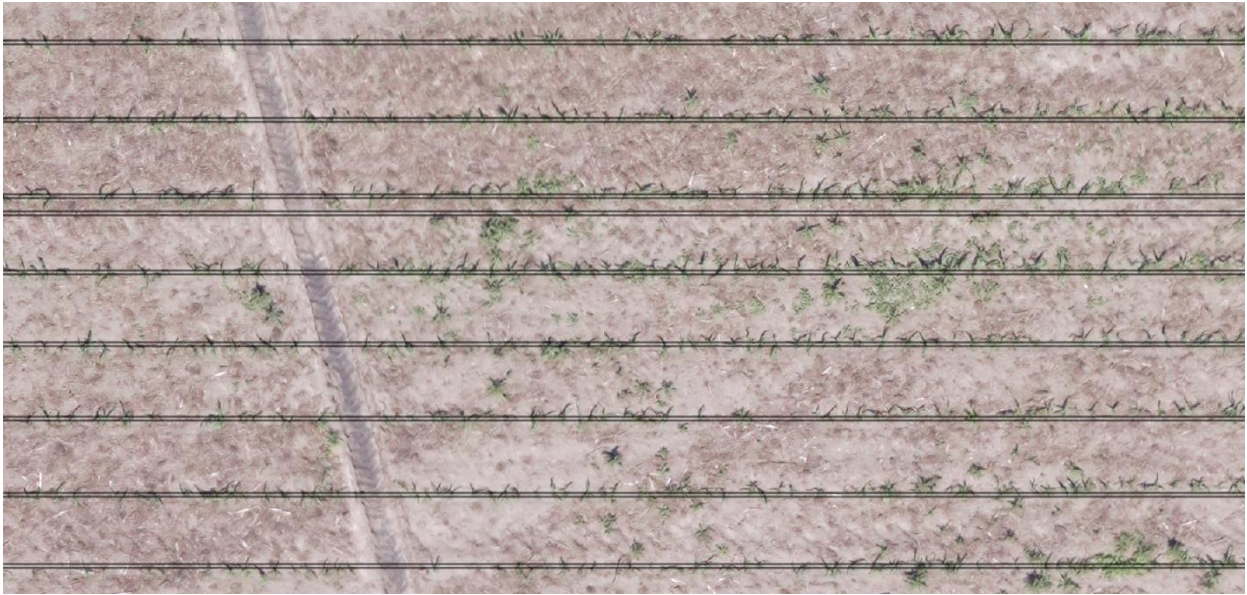

**Figure 13. Sample image where the proposed PIP algorithm detected two lines over a single corn row on the ExGI imagery of a corn field. Imagery captured by Sony Alpha 7R II, flown at 350 ft AGL.**

Most of the published studies have made use of Hough Line Transform (HLT) to identify crop rows (Bakker et al., 2008; Chen et al., 2021; Leemans & Destain, 2006; Marchant & Brivot, 1995; Soares et al., 2018a, 2018b). The HLT requires edge detection step prior to generate line patterns (Mukhopadhyay & Chaudhuri, 2015). The HLT method was applied to this study and found to be ineffective for imagery covering such large area. Some recent studies have used deep learning methods to identify crop rows for weed detection (Bah et al., 2018; Czymmek et al., 2019; R. de Silva et al., 2021; Su et al., 2021). This method requires initial training of models with image data which is time-consuming and requires a lot of training datasets.

The PIP algorithm proposed in this research has an advantage over those methods because there is no requirement for training a model for row identification. In addition, most of the existing research has used RGB imagery directly for identifying plant rows (Guerrero et al., 2013; Hassanein et al., 2019; Romeo et al., 2012). The direct use of RGB imagery in any computer vision and machine learning module requires classification of the imagery into binary format, because the machine learning modules and computer vision algorithm work by processing binary information of the image (Patrignani & Ochsner, 2015; Riehle et al., 2020; Saxena & Armstrong, 2014). Instead of using direct RGB imagery, which requires classification and feature extraction, we have made use of segmented excess green binary imagery as an input for the algorithm. As a result, the processing time of the algorithm is quite fast. In our case, it detected the corn rows of six SSWC plots in less than 4 seconds, which is one of the greatest achievements of the algorithm.

The area covered by the six SSWC test plots was approximately 7.6 acres, however, the algorithm was used to calculate the lines over the whole orthomosaic to evaluate the processing time of the algorithm for a larger acreage. The algorithm created lines of corn rows over approximately 42 acres of corn in less than 6 seconds. This processing speed may vary depending on the processing power of the computer. From these two results, we can roughly estimate that the algorithm would create corn row lines over 160 acres (or a quarter-section of land) in less than 20 seconds. The author was unable to find supporting literature that has processed this much amount of land imagery for crop row detection. Most of the literature stated in this study used machine learning and deep learning models to predict crop rows, which is time consuming compared to the proposed approach because all machine learning and deep learning models require the training of a large dataset. That is a very important aspect as one might consider the applications of this approach for SSWC on a larger scale.

Moreover, if we could get the row location from equipment such as a planter, the implementation of this approach would be much faster. There is a commercial planter from John Deere with a technology called Autopath, which provides more accurate row guidance and can records the row information during the planting of seed (John Deere, 2020).

## 4.2. Weed Mapping Accuracy

As described on item 3.7, AgSMS software was used to convert the prescription layer created in ArcGIS Pro to a format that the sprayer could understand and act upon it. Once imported into AgSMS, the prescription map was exported using the "full device setup to a field display/monitor" option, so a prescription map would be created in a format supported by the Raven Viper4+ cab computer. The map generation seemed to work just fine, and the field application continued on. It was not until later, when working on a study related to this one, when members of Dr. Flores' lab noticed that there was a difference between a prescription layer created in ArcGIS Pro and the output from AgSMS software when that layer was imported and then exported as described above. Upon further investigation, it was found that the same issue had occurred to this field study, resulting on very different prescription maps first intended for the study (Figures14a and 14b). Although a workaround solution for the problem was identified, due to the nature of this study, the research team had only one chance to get the application right, which was not possible due to those unforeseen changes to Rx map made by AgSMS. Those (random) changes led to the loss of the accurate spatial distribution of the cells free of weeds across the field. Figures 14a and 14b show a mismatch between the area that originally should not be sprayed (ArcGIS Pro file) and the areas that were actually not sprayed (AgSMS file) for SSWC-1 test plot.

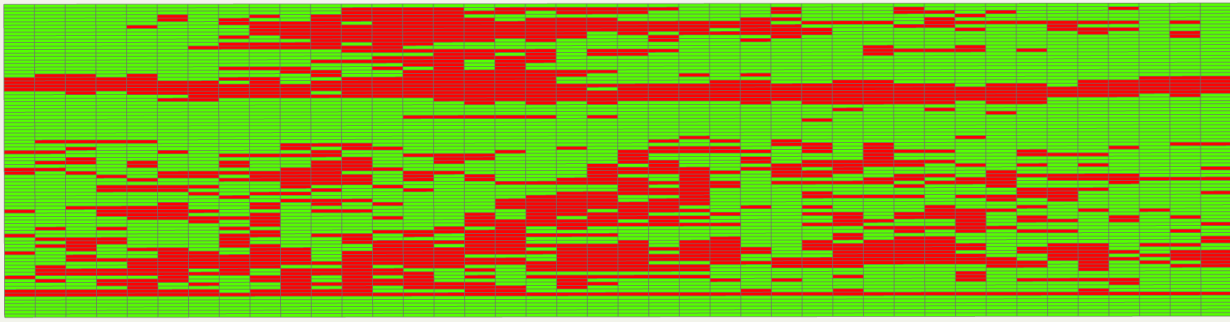

a)

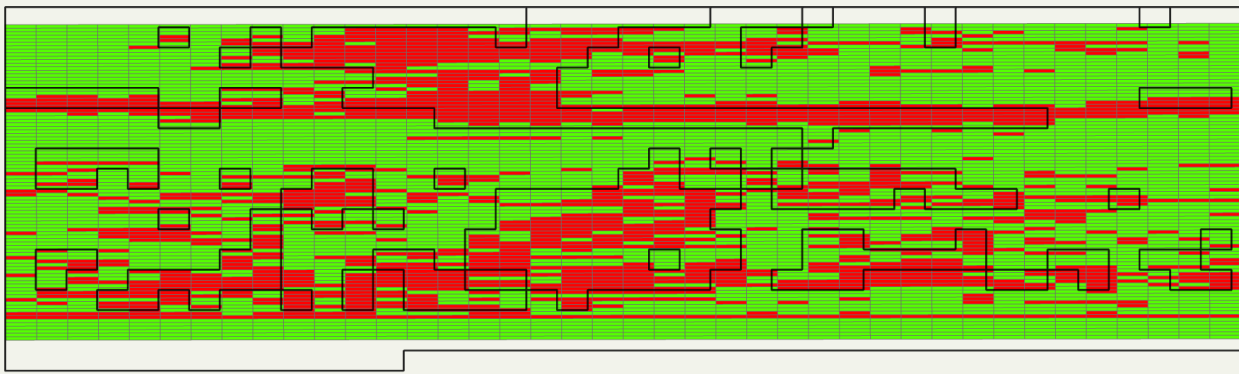

b)

**Figure 14. a) Prescription map created in the ArcGIS Pro for SSWC-1 where all red grids were set as no-spray and all green grids were set as spray regions on the imagery of a corn field; b) Overlap of ArcGIS map with the AgSMS no-spray regions (regions enclosed within dark black lines); the red regions outside of the AgSMS no-spray regions was not sprayed.**

To overcome that issue and to try get some useful results from the field herbicide application, both the original prescription map (created in ArcGIS Pro) and the as-applied were overlapped to better understand the magnitude of changes made by AgSMS when compared to the intended application map, focusing mainly on the cells with as 0 gal/ac rate (free of weeds). Those results are shown on Table 2. The area of no-spray regions set on the ArcGIS map (red grids in Figure 14 a), and that no-spray area included inside the no-spray region of AgSMS map (all highlighted grids in Figure 14 b) is shown below for all six SSWC test plots (Table2).

**Table 2. Impact, in terms of area loss, of the unintended AgSMS software changes to the prescription map in area that was not supposed to be sprayed based on the original (ArcGIS Pro) prescription map.**

| Plot | ArcGIS Pro map, $m^2$ | AgSMS map, $m^2$ | Area loss due to AgSMS changes, $m^2$ | Area loss due to AgSMS changes, % |
|---|---|---|---|---|
| SSWC-1 | 2598.6 | 1600.1 | 998.6 | 38.4 |
| SSWC-2 | 3316.9 | 2209.1 | 1107.8 | 33.4 |
| SSWC-3 | 2302.3 | 1146.1 | 1156.3 | 50.2 |
| SSWC-4 | 1739.2 | 841.8 | 897.4 | 51.6 |
| SSWC-5 | 2241.8 | 1280.7 | 961.2 | 42.9 |
| SSWC-6 | 1757.8 | 841.8 | 915.9 | 52.1 |
| **TOTAL** | **13956.7** | **7919.5** | **6037.2** | **43.3** |

From the data on Table 2, one can notice that 43.3% of the original no-spray region (from ArcGIS map) for the SSWC treatment did not receive the intended application. The observation that AgSMS makes those changes to make things "easier" to the controllers on variable rate equipment seems to be supported by the values found across the replications for the SSWC treatment. Data on Table 2 shows that changes were smaller on plots where there were large areas of no-sprayed cell grouped together (forming a larger polygon), which makes things "easier" on the sprayer variable rate controller, so not many changes were needed. On the other hand, plots with originally a large number of individual no-spray cells (SSWC 3, 4, and 6) would make things "harder" on the sprayer, so AgSMS made more changes to those plots to make things "easier" on the controller.

The total area that was set as no-spray in the prescription map created on ArcGIS Pro was 13,956.7 $m^2$, but the AgSMS software reduced the total no-spray regions to 10,098.2 $m^2$. In addition, out of that only 7,917.5 $m^2$ were free of weeds, which is 78.0% of the total no-spray region in the AgSMS map. Thus, the remaining 22.0% of area was infested with weeds, but the changes made by AgSMS caused those areas to be treated as a no-spray area. That becomes a big issue when trying to evaluate the efficacy of SSWC approach proposed here. The premise for the proposed approach to be successful is that the weed pressure on the no-spray cells after harvest is equal or less than those cells that were sprayed in the spring. Since the AgSMS changes caused the sprayer not spraying cells with weeds in the spring, it is challenging to make a clear assessment of the differences between the treatments. Moreover, when the original prescription map was created in ArcGIS Pro, the weeds vegetation layer selected was completely within or within the cells. This led to not consider some small fraction of weeds that were not completely inside the boundary of cells.

### 4.3. As-applied Map Accuracy

The accuracy of as-applied map was calculated based on no-spray regions that were set in AgSMS map. Figure 15 shows the comparison between the Rx map and the as-applied map. All the regions enclosed by black lines were the no-spray regions set in the Rx map. Purple shaded regions inside the no-spray cells of the Rx map are the no-spray regions recorded in the as-applied map, while all the regions except the purple regions were sprayed.

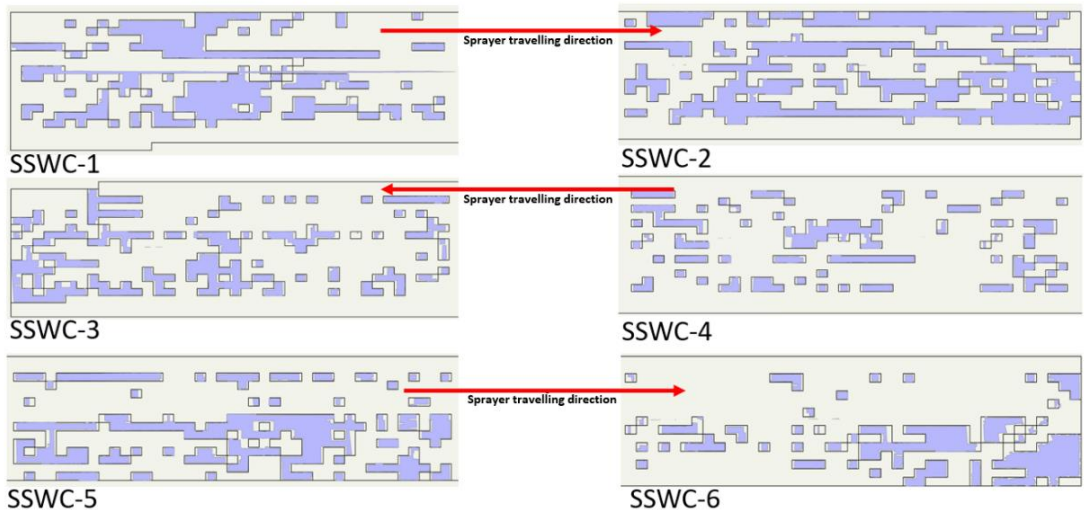

**Figure 15. Overlap of Rx map, generated by AgSMS, where the regions enclosed by dark black lines where the no-spray regions (0 gal/ac), and the as-applied map. Purple color regions are the area where the sprayer turned its nozzles off while operating at 6.5 mph speed (approximate), while all the remaining area was sprayed (15 gal/ac).**

From the data in Figure 15, it was determined that the sprayer was able to turn off its nozzles over 7,919.5 m$^2$, from a total of 10,098.1 m$^2$, across all six SSWC replicates. Those results show that the sprayer was able to perform no-spray only on 78.4 % of the no-spray. The relative error was calculated at = -0.216, which means the measured value (from the as-applied map) was 21.6% less than the expected value (from AgSMS map). Table 3 shows the detailed values of no-spray areas which was set in the AgSMS Rx map for the six SSWC treatment plots and the no-spray area that was recorded in the as-applied map.

**Table 3. Comparison of no-spray regions that was set by using AgSMS software (column 2) with the no-spray regions that was recorded in the as-applied map (column 3) downloaded from the cab computer of Case IH Patriot 4440 sprayer.**

| Test-Plot | AgSMS map no-spray area, m$^2$ | As-applied map no-spray, m$^2$ | Sprayed area in no-spray regions, m$^2$ (as-applied map) | Sprayed area % in no-spray regions (as-applied map) |
|---|---|---|---|---|
| SSWC-1 | 1822.3 | 1600.1 | 222.2 | 12.2 |
| SSWC-2 | 2648.9 | 2209.1 | 439.8 | 16.6 |
| SSWC-3 | 1591.2 | 1146.1 | 445.1 | 27.9 |
| SSWC-4 | 1182.3 | 841.8 | 340.5 | 28.8 |
| SSWC-5 | 1688.9 | 1280.7 | 408.3 | 24.2 |
| SSWC-6 | 1164.5 | 841.8 | 322.7 | 27.7 |
| **TOTAL** | **10098.1** | **7919.5** | **2178.6** | **21.5** |

The reduction of the no-spray area observed in the as-applied map does not seem to follow a specific pattern across the field, which creates difficulty in understanding the factor(s) driving those results. However, the majority of the reduction in the no-spray area was noticed at the edges of no-spray regions set in Rx map. From the visual comparison of as-applied map with the Rx map, majority of as applied map showed that the sprayer's nozzle stopped spraying only after reaching the no-spray zone. Most of the edges that were sprayed inside the no-spray regions showed that the nozzles did not start spraying instantly as the no-spray regions started. One of the possible explanations for such behavior might a combination of the sprayer's travelling direction and speed, where the sprayer is moving too fast to stop spraying right on the edge of cells.

Moreover, it was noticed that the sprayer missed spraying some small no-spray regions that were surrounded by spray regions in the Rx prescription map.

The 21.5% of land area which was sprayed inside no-spray region might have a direct impact on the economic aspects of this approach. One the goals of this SSWC approach is to reduce the acreage requiring herbicide application. In the case of this study, this approach resulted in a reduction of 35% on herbicide application. At the level of herbicide savings it might make economic sense for such SSWC approach to be adopted, but once we add the 21.5% error on the no-sprayed area, that scenario might change and the whole approach might no longer be attractive to farmers.

## 4.4. Effectiveness of the SSWC approach

The two treatments, SSWC and NO-SSWC, were significantly different in terms of weed area ($F_{1,5}$=11.41, p<0.0197). The amount of weeds present in SSWC treatment plots was 3.4 times higher than the amount of weeds that were present in SSWC plots. The estimated sum of the mean value for the area of weeds in six SSWC treatment plots was 87.02 $m^2$, which was higher (3.4 times) than the estimated sum of the mean value for the area of weeds in no-SSWC treatment plots (25.5 $m^2$). Figure 16 shows the graphical representation of weed area in SSWM and NO-SSWM treatment plots.

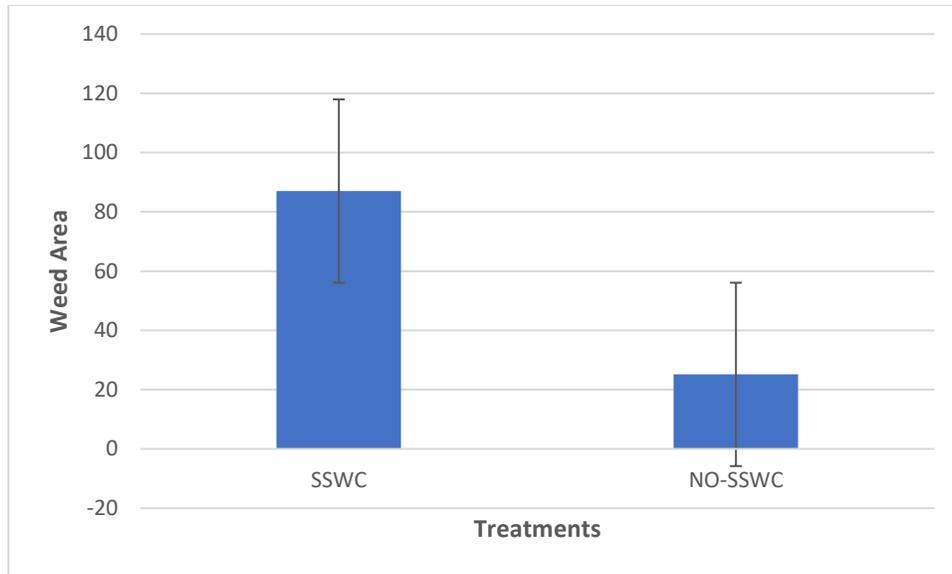

**Figure 16. Weed area estimated in SSWC and NO-SSWC treatment plots from the post-harvest imagery, calculated from SAS PROC MIXED using REML mixed procedure**

One of the possible answers for why the SSWC test plots have higher weeds is that the Rx map integrated with sprayer had almost 22% of the area that was infested with weeds inside the no-spray region of AgSMS (as explained in section 4.2, and figure 14 b), and the application was not able to spray herbicides over that 22% weed infested regions. Because of this, the presence of weeds in the SSWC test plots was 3.4 times higher than the no-SSWC test plots in the post-harvest imagery. The findings from this study could be a foundation to pursue various research questions that our study could not address, because of time and other limiting factors. There might be other several possible reasons for the number of weeds in the SSWC treatment plots to be higher than the NO-SSWC treatment plots. Some of those reasons are addressed below.

Although the imagery obtained from the field had high resolution (GSD= 0.63 cm per pixel), we might have missed smaller weeds during the data collection of the early growing season. Those weed detection skips might be one of the reasons why we observed weeds in the SSWC plots after the corn harvest. One could argue that image resolution should be increased to avoid those skips, but there are other challenges to achieve that, such as cost of better hardware (UASs and sensors), time to collect data (by flying lower with the same equipment used on this study), and time to process the data (lower altitude flight leads to many more pictures captured, which increases processing time).

Since the cell size was made to fit for each nozzle of the sprayer, there is a chance that the sprayer missed spraying some weeds that were within corn rows. The width of each grid was 0.5 m [1.67 feet] which was exactly equal to the spacing between each nozzle in the sprayer's boom, but that is less than the corn row spacing (0.7 m [2.5 ft]).

This year was considered a bad year for implementing this research in terms of calculating the effectiveness of research because of the high weed pressure at the time of the post-emergence application. According to information collected from the field crew at the Carrington Research Extension Center, there was a high probability that pre-emergence treatment did not work effectively because of weather conditions.

Another possible reason for having more weeds in SSWC plots might be because we applied the post-emergence treatment too late, and the treatment was as not effective as anticipated because of the size of some weeds. That is supported by the imagery collected after the corn harvest, where there was noticeably high weed pressure, even in those areas that received a blanket application of herbicide. Some of the weeds might have been too advanced in their growth stage to be controlled with the chemical rates applied.

# 5. CONCLUSION

Weed management is vital to agriculture and UAS imagery can be a very useful tool to implement SSWC in corn fields. This study showed case for using UAS imagery to create an accurate weed prescription map and integrate that map with a commercial size sprayer.

In this study we presented a novel approach of projecting pixel intensities for identifying corn rows, which plays a crucial role on generating a weed map from the field using our proposed approach. Once the corn rows are identified, they were removed from the imagery and the remaining vegetation as classified as weeds. While other machine learning models and computer-vision approach rely on Hough Line Transform and training of a lot of images for row detection, our "PIP algorithm" directly estimated lines over each corn row when the binary excess green imagery was provided as an input.

Our approach of identifying corn rows first and removing all the corn rows from the UAS imagery is an effective way to identify the weeds that grows during the early growing season of corn plants. In addition, our approach of mapping weeds based on grid cells is promising way of creating weed prescription map for SSWC. Using the prescription map, the workflow proved a promising capability in terms of individual nozzle control.

# 6. SUGGESTIONS FOR FUTURE RESEARCH

Since the number of weeds that were present in SSWC plots was higher than NO-SSWC plots in post-harvest imagery, we have several future recommendations that could potentially increase the efficacy of the approach.

The grids that were generated for the weed prescription map in this study were very small in terms of width (i.e. 1.6×10 ft). Each cell laid under one of the sprayer's nozzles, which could have reduced the effectiveness of the entire research approach. Alternative map grids with approximately 10×10 ft cell size would have been useful for better efficacy. Therefore, a comparison of bigger cell size prescription maps must be conducted in further research. Furthermore, we suggest that the sprayer would have done a poor job if the sprayer speed was operated at 12 mph or more. Since this study was implemented using a sprayer set at approx. 6.5 mph, a study of this approach at a higher sprayer speed would be very useful in the future.

Moreover, the research was conducted on approximately 42 acres of land. In order to make this approach applicable for farming, larger land coverage would be very useful. The scalability of this study needs to increase in terms of land coverage area. In addition, the image collection time during this research was around 52 minutes (early growing season), which may be unfavorable in terms of time value. Identification of new hardware and software for reducing data collection time is required for future improvement of this approach.

Since most of the work in this study was done by human intervention, it would be very beneficial if the entire process of row detection to weed mapping could be automated. Real-time weed mapping using this approach would be very useful in the future of weed management.